\pdfoutput=1

\documentclass[11pt]{article}

\usepackage[final]{emnlp22/ACL2023}

\usepackage{times}
\usepackage{latexsym}

\usepackage[T1]{fontenc}

\usepackage[utf8]{inputenc}

\usepackage{microtype}

\usepackage{inconsolata}

\usepackage{amsmath,amsfonts,bm}

\newcommand{\Ni}{({\em i})~}
\newcommand{\Nii}{({\em ii})~}

\def\eqref#1{equation~\ref{#1}}

\def\1{\bm{1}}

\def\vh{{\bm{h}}}

\def\vo{{\bm{o}}}

\def\vx{{\bm{x}}}

\def\vz{{\bm{z}}}

\DeclareMathAlphabet{\mathsfit}{\encodingdefault}{\sfdefault}{m}{sl}
\SetMathAlphabet{\mathsfit}{bold}{\encodingdefault}{\sfdefault}{bx}{n}

\def\gG{{\mathcal{G}}}

\def\gL{{\mathcal{L}}}

\usepackage{hyperref}
\usepackage{url}

\usepackage{multirow}
\usepackage{makecell}
\usepackage{graphicx}
\usepackage{tablefootnote}
\usepackage{adjustbox}
\usepackage{subfiles}
\usepackage{booktabs}
\usepackage{cleveref}
\usepackage{wrapfig}
\usepackage{todonotes}
\usepackage{enumitem}

\usepackage{subcaption}
\usepackage{latexsym}
\usepackage{color, colortbl}
\usepackage{amsmath}
\usepackage{amssymb}

\crefformat{section}{\S#2#1#3}
\crefformat{subsection}{\S#2#1#3}

\crefformat{subsubsection}{\S#2#1#3}

\title{Towards Robust Low-Resource Fine-Tuning with \\ Multi-View Compressed Representations}

\author{
Linlin Liu$^{1,2}$\thanks{\; Equal contribution, order decided by coin flip. Linlin Liu and Xingxuan Li are under the Joint Ph.D. Program between Alibaba and Nanyang Technological University.} ~~ 
Xingxuan Li$^{1,2}$\footnotemark[1] ~~ 
Megh Thakkar$^2$ ~
Xin Li$^1$ ~
\\
\textbf{Shafiq Joty}$^{2,3}$ ~
\textbf{Luo Si}$^1$ ~
\textbf{Lidong Bing}$^1$\thanks{\; Corresponding author.}\\
$^1$DAMO Academy, Alibaba Group ~ \\
$^2$Nanyang Technological University~ $^3$Salesforce AI\\
\{linlin.liu, xingxuan.li, xinting.lx, luo.si, l.bing\}@alibaba-inc.com\\
megh.1211@gmail.com ~~~ srjoty@ntu.edu.sg
}

\begin{document}
\maketitle

\begin{abstract}

Due to the huge amount of parameters, fine-tuning of pretrained language models (PLMs) is prone to overfitting in the low resource scenarios. In this work, we present a novel method that operates on the hidden representations of a PLM to reduce overfitting. During fine-tuning, our method inserts random autoencoders between the hidden layers of a PLM, which transform activations from the previous layers into multi-view compressed representations before feeding them into the upper layers. The autoencoders are plugged out after fine-tuning, so our method does not add extra parameters or increase computation cost during inference. Our method demonstrates promising performance improvement across a wide range of sequence- and token-level low-resource NLP tasks.
{Our code is available at \href{https://github.com/DAMO-NLP-SG/MVCR}{https://github.com/DAMO-NLP-SG/MVCR}}.

\end{abstract}
\section{Introduction}
Fine-tuning pretrained language models (PLMs) \citep{devlin-etal-2019-bert,conneau2019cross,liu-etal-2020-multilingual-denoising} provides an efficient way to transfer knowledge gained from large scale text corpora to downstream NLP tasks, which has achieved state-of-the-art performance on a wide range of tasks \citep{NEURIPS2019_dc6a7e65,yamada-etal-2020-luke,chi-etal-2021-infoxlm,sun2021ernie,zhang-etal-2021-towards-generative,zhang-etal-2021-aspect-sentiment,chia-etal-2022-relationprompt,tan-etal-2022-document,zhou-etal-2022-conner}. However, most of the PLMs are designed for general purpose representation learning \citep{roberta,conneau-etal-2020-unsupervised}, so the learned representations unavoidably contain abundant features irrelevant to the downstream tasks. Moreover, the PLMs typically possess  a huge amount of parameters (often 100+ millions) \citep{NEURIPS2020_1457c0d6,min2021recent}, which makes them more expressive compared with simple models, and hence more vulnerable to overfitting noise or irrelevant features during fine-tuning, especially in the low-resource scenarios.

There has been a long line of research on devising methods to prevent large neural models from overfitting. The most common ones can be roughly grouped into three main categories: data augmentation \citep{devries2017improved,ding-etal-2020-daga,liu-etal-2021-mulda,feng-etal-2021-survey, xu2022peerda, zhou-etal-2022-melm}, parameter/activation regularization \citep{weightdecay,dropout,mixout,li2020unsupervised} and label smoothing \citep{szegedy2016rethinking,Yuan_2020_CVPR}, which, from bottom to top, operates on data samples, model parameters/activations and data labels, respectively. 

Data augmentation methods, such as back-translation \citep{sennrich-etal-2016-improving} and masked prediction \cite{bari-etal-2021-uxla}, are usually designed based on our prior knowledge about the data. Though simple, many of them have proved to be quite effective. Activation (hidden representation) regularization methods are typically orthogonal to other methods and can be used together to improve model robustness from different aspects. However, since the neural models are often treated as a black box, the features encoded in hidden layers are often less interpretable. Therefore, it is more challenging to apply similar augmentation techniques to the hidden representations of a neural model.


Prior studies \citep{yosinski2015understanding,allen2020towards,fu2022effectiveness} observe that neural models trained with different regularization or initialization can capture different features of the same input for prediction. Inspired by this finding, in this work we propose a novel method for hidden representation augmentation. Specifically, we insert a set of randomly initialized autoencoders (AEs) \citep{rumelhart1985learning,baldi2012autoencoders} between the layers of a PLM, and use them to capture different features from the original representations, and then transform them into \textbf{M}ulti-\textbf{V}iew \textbf{C}ompressed \textbf{R}epresentations (\textbf{MVCR}) to improve robustness during fine-tuning on target tasks. Given a hidden representation, an AE  first encodes it into a compressed representation of smaller dimension $d$, and then decodes it back to the original dimension. The compressed representation can capture the main variance of the data. 
Therefore, with a set of AEs of varying $d$, if we select a random one to transform the hidden representations in each fine-tuning step, the same or similar input is compressed with varying compression dimensions.
{And the upper-level PLM layers will be fed with more diverse and compressed representations for learning, which illustrates the ``multi-view'' concept.}
We also propose a tailored hierarchical AE to further increase representation diversity. 
Crucially, after fine-tuning the PLM with the AEs, the AEs can be plugged out, so they do not add any extra parameters or computation during inference.

\begin{figure}[t!]
\centering
{\footnotesize
\includegraphics[width=0.16\columnwidth]{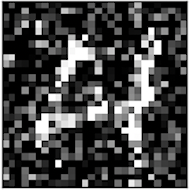}
\includegraphics[width=0.16\columnwidth]{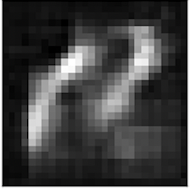}
\includegraphics[width=0.16\columnwidth]{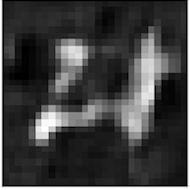}
\includegraphics[width=0.16\columnwidth]{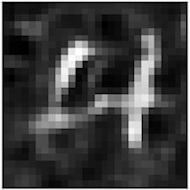}\\
\includegraphics[width=0.16\columnwidth]{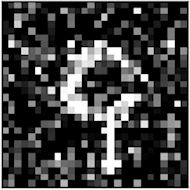}
\includegraphics[width=0.16\columnwidth]{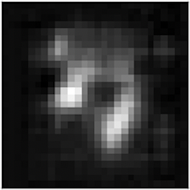}
\includegraphics[width=0.16\columnwidth]{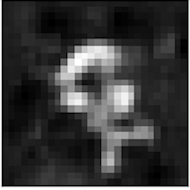}
\includegraphics[width=0.16\columnwidth]{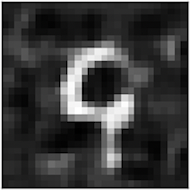}\\
\includegraphics[width=0.16\columnwidth]{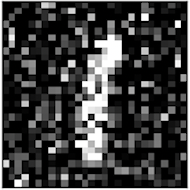}
\includegraphics[width=0.16\columnwidth]{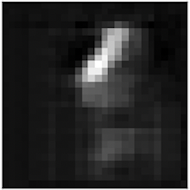}
\includegraphics[width=0.16\columnwidth]{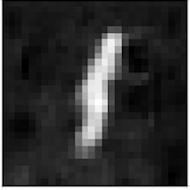}
\includegraphics[width=0.16\columnwidth]{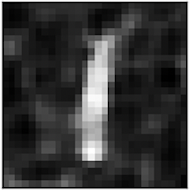}\\
\includegraphics[width=0.16\columnwidth]{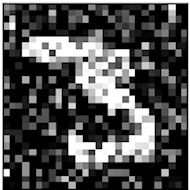}
\includegraphics[width=0.16\columnwidth]{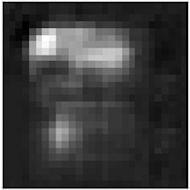}
\includegraphics[width=0.16\columnwidth]{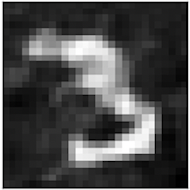}
\includegraphics[width=0.16\columnwidth]{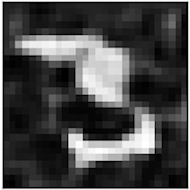}\\
\includegraphics[width=0.16\columnwidth]{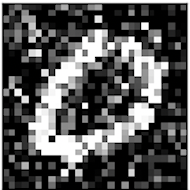}
\includegraphics[width=0.16\columnwidth]{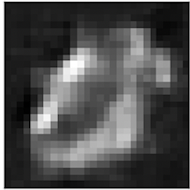}
\includegraphics[width=0.16\columnwidth]{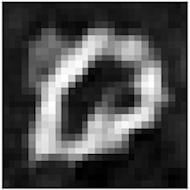}
\includegraphics[width=0.16\columnwidth]{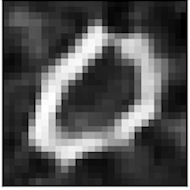}\\
\makebox[0.16\columnwidth]{(a)}
\makebox[0.16\columnwidth]{(b)}
\makebox[0.16\columnwidth]{(c)}
\makebox[0.16\columnwidth]{(d)} \\
}
\caption{An illustration of the results reconstructed by autoencoders (AEs) with varying compression ratios: (a) are the inputs (28×28 pixel) to the AEs, and (b) - (d) are the reconstructed images from compressed representations of dimensions 49, 98 and 392, respectively.} 
\label{fig:autoencoder_mnist}
\end{figure}


We have designed a toy experiment to help illustrate our idea in a more intuitive way. We add uniform random Gaussian noise to the MNIST \citep{lecun1998gradient} digits and train autoencoders with different compression ratios to reconstruct the noisy input images.
As shown in Fig.~\ref{fig:autoencoder_mnist}, the compression dimension $d$ controls the amount of information to be preserved in the latent  space. With a small $d$, the AE removes most of the background noise and preserves mostly the crucial shape information about the digits. In Fig.~\ref{fig:autoencoder_mnist}b, part of the shape information is also discarded due to high compression ratio. In the extreme case, when $d=1$, the AE would discard almost all of the information from input. Thus, when a small $d$ is used to compress a PLM's hidden representations during fine-tuning, the reconstructed representations will help: \Ni reduce overfitting to noise or task irrelevant features since the high compression ratio can help remove noise; \Nii force the PLM to utilize different relevant features to prevent it from becoming overconfident about a certain subset of them. Since the AE-reconstructed representation may preserve little information when $d$ is small (Fig.~\ref{fig:autoencoder_mnist}b), the upper layers are forced to extract only relevant features from the limited information. Besides, the shortcut learning problem \citep{geirhos2020shortcut} is often caused by learning features that fail to generalize, for example relying on the grasses in background to predict sheep. Our method may force the model to use the compressed representation without grass features. Therefore, it is potentially helpful to mitigate shortcut learning. As shown in Fig. \ref{fig:autoencoder_mnist}(d), with a larger $d$, most information about the digits can be reconstructed, and noises also start to appear in the background. Hence, AEs with varying $d$ can transform a PLM's hidden representations into different \emph{views} to increase diversity.


We conduct extensive experiments to verify the effectiveness of MVCR. Compared to many strong baseline methods, MVCR demonstrates consistent performance improvement across a wide range of sequence- and token-level tasks in low-resource adaptation scenarios. We also present abundant ablation studies to justify the design. In summary, our main contributions are:

\begin{itemize}[leftmargin=*,topsep=2pt,itemsep=2pt,parsep=0pt]
    \item Propose a novel method to improve low-resource fine-tuning, which leverages AEs to transform representations of a PLM into multi-view compressed representations to reduce overfitting.
    \item Design an effective hierarchical variant of the AE to introduce more diversity in fine-tuning.
    \item Present a plug-in and plug-out fine-tuning approach tailored for our method, which does not add extra parameter or computation during inference. 
    \item Conduct extensive experiments to verify the effectiveness of our method, and run ablation studies to justify our design.
\end{itemize}

\section{Related Work}

Overfitting is a long-standing problem in large neural model training, which has attracted broad interest from the research communities \citep{sarle1996stopped,hawkins2004problem,salman2019overfitting,santos2022avoiding,hu2023llmadapters,he-etal-2021-effectiveness}. To better capture the massive information from large-scale text corpora during pretraining, the PLMs are often over-parameterized, which makes them prone to overfitting. The commonly used methods to reduce overfitting can be grouped into three main categories: data augmentation, parameter regularization, and label smoothing.

Data augmentation methods \citep{devries2017improved,liu-etal-2021-mulda,feng-etal-2021-survey} are usually applied to increase training sample diversity. Most of the widely used methods, such as synonym
replacement \citep{karimi-etal-2021-aeda-easier}, masked prediction \cite{bari-etal-2021-uxla} and  back-translation \citep{sennrich-etal-2016-improving}, fall into this category. Label smoothing methods \citep{szegedy2016rethinking,Yuan_2020_CVPR} are applied to the data labels to prevent overconfidence and to encourage smoother decision boundaries. Some hybrid methods, like MixUp \citep{zhang2018mixup,verma2019manifold}, are proposed to manipulate both data sample and label, or hidden representation and label. Most parameter/hidden representation regularization methods are orthogonal to the methods discussed above, so they can be used as effective complements.

Neural models are often treated as a black box, so it is more challenging to design efficient parameter or hidden representation regularization methods. Existing methods mainly focus on reducing model expressiveness or adding noise to hidden representations. Weight decay \citep{weightdecay} enforces $L_{2}$ norm of the parameters to reduce model expressiveness. Dropout \cite{dropout} randomly replaces elements in hidden representations with $0$, which is believed to add more diversity and prevent overconfidence on certain features. 
Inspired by dropout, Mixout \citep{mixout} stochastically mixes the current and initial model parameters during training. \citet{vibert} leverage variational information bottleneck (VIB) \citep{alemi2016deep} to help models to learn more concise and task relevant features. However, VIB is limited to regularize last layer sequence-level representations only, while our method can be applied to any layer, and also supports token-level tasks like NER and POS tagging.


\begin{figure*}[t!]
    \centering
    \includegraphics[width=0.85\textwidth]{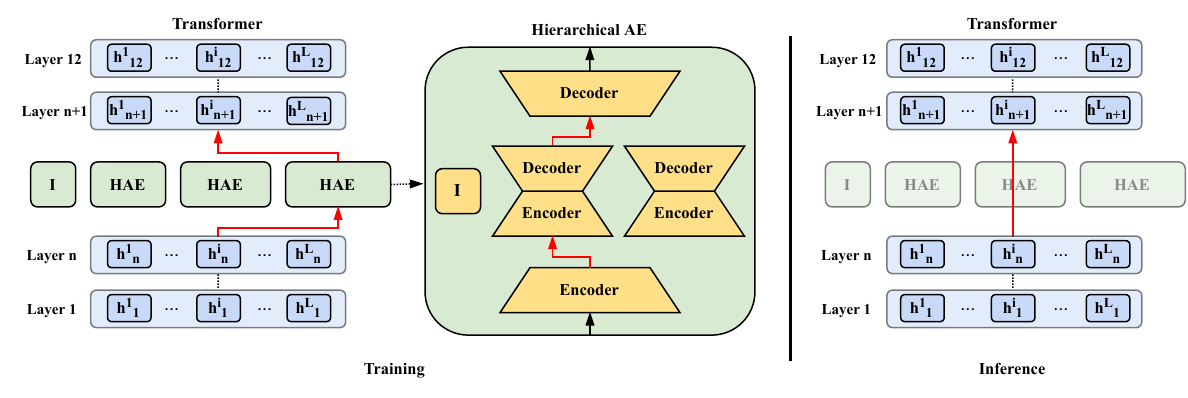}
    \caption{Illustration of token-level Multi-View Compressed Representations (MVCR) with three stochastic hierarchical autoencoders (HAEs) inserted between the transformer layers. During training, the output of layer $n$ is either passed through a randomly selected HAE or directly passed to layer $n+1$ (denoted as ``I'' in the figure). If an HAE is picked, the output of the outer encoder is either passed through a randomly selected sub-AE or directly passed to the outer decoder (via ``I''). In inference, we drop MVCR without adding parameters to the original PLM.}
    \label{fig:model}
\end{figure*}

\section{Methodology}


We first formulate using neural networks over the hidden representation of different layers of deep learning  architectures as effective augmentation modules (\cref{subsection:compression}) and then devise a novel hierarchical autodencoder (HAE) to increase stochasticity and diversity (\cref{subsection:HAE}). We utilise our novel HAE as a compression module for hidden representations within PLMs, and introduce our method Multi-View Compressed Representation (MVCR) (\cref{subsection:MVCR}). We finally discuss the training, inference, and optimization of MVCR. An overview of MVCR is presented in Fig.~\ref{fig:model}.

\subsection{Generalizing Neural Networks as Effective Augmentation Modules}\label{subsection:compression}

Data augmentation \cite{Simard1998} aims at increasing the diversity of training data while preserving quality for more generalized training \cite{shorten2019survey}. It can be formalized by the Vicinal Risk Minimization principle \cite{VRM_NIPS2000}, which aims to enlarge the support of the training distribution by generating {new} data points from a \emph{vicinity distribution} around each training example \cite{zhang2018mixup}.
We conjecture that shallow neural networks can be used between the layers of large-scale PLMs to construct such vicinity distribution in the latent space to facilitate diversity and generalizability.

We consider a neural network $g(\cdot)$, and denote its forward pass $F(\cdot)$  for an input $\vx$ with $F(\vx)=g(\vx)$.
We denote a set of $M$ such networks with $\mathcal{G}=\{g_{1}(\cdot),  \dots, g_{M}(\cdot) \}$, where each candidate network $g_{i}(\cdot)$ outputs a different \textit{"view"} of a given input. We treat $\gG$ as a stochastic network and define a stochastic forward pass $F^S(\cdot, \mathcal{G})$, where a candidate network $g_{i}(\cdot)$ is randomly chosen from the pool $\mathcal{G}$ in each step, enabling diversity due to different non-linear transformations. Formally, for an input $\vx$, we obtain the output $\vo$ of network $g_{i}(\cdot)$ using $F^S$ as,
\begin{equation}
    \vo = F^S(\vx, \mathcal{G}) = g_{i}(\vx), \quad i \in \{1,\dots, M\} 
    \label{eqn:SF}
\end{equation}
For a chosen candidate network $g_{i}$, the output $\vo$ represents a network dependent ``\textit{view}'' of input $\vx$.

We now formalize using $g(\cdot)$ over the hidden representation of large-scale PLMs for effective generalization. Let $f(\cdot)$ denote \textit{any} general transformer-based PLM  containing $N$ hidden layers with $f_{n}(\cdot)$ being the $n$-th layer and $\vh_n$ being the activations at that layer for $n \in \{1,\dots, N\}$, and let $\vh_0$ denote the input embeddings.
We consider a set of layers $\gL \subset \{1,\dots, N\}$ where we insert our stochastic network $\mathcal{G}$ for augmenting the hidden representations during the forward pass. To this end, we substitute the standard forward pass of the PLM $f(\cdot)$ during the \textit{training} phase with the stochastic forward pass $F^S(\cdot, \mathcal{G})$. Formally,
\begin{equation}\small
    \vh_n = \begin{cases}
            F^S(f_{n}(\vh_{n-1}), \mathcal{G}_n), & n \in \mathcal{L} \\
            f_{n}(\vh_{n-1}), & n \notin \mathcal{L}
          \end{cases}
  \label{eqn:hn}
\end{equation}
We now devise a novel hierarchical autoencoder network which can be effectively used for augmenting hidden representations of large-scale PLMs.


\subsection{Stochastic Hierarchical Autoencoder}\label{subsection:HAE}

Autoencoders (AEs) are a special kind of neural networks where the input dimension is the same as the output dimension \cite{rumelhart1985learning}. For an input $\vx \in \mathbb{R}^d$, we define a simple autoencoder $\mathrm{AE}_{d, \hat{d}}(\cdot)$ with compression dimension $\hat{d}$ as a sequential combination of a  feed-forward down-projection layer $D_{d,\hat{d}} (\cdot)$ and an up-projection layer $U_{\hat{d}, d} (\cdot)$. Given input $\vx$, the output $\vo \in \mathbb{R}^d$ of the autoencoder can be represented as:
\begin{equation}\small
    \vo = \mathrm{AE}_{d, \hat{d}}(\vx) = U(D(\vx))
    \label{eqn:ae_fw}
\end{equation}

\paragraph{Hierarchical Autoencoder (HAE)} We extend a standard autoencoder to a hierarchical autoencoder and present its overview in Fig. \ref{fig:model} (Hierarchical AE). 
We do this by inserting a sub-autoencoder $\mathrm{AE}_{\hat{d}, d'}(\cdot)$ with compression dimension $d'$ within an autoencoder $\mathrm{AE}_{d, \hat{d}}(\cdot)$, where $d' < \hat{d}$. For $\mathrm{AE}_{\hat{d}, d'}(\cdot)$, the down- and up-projection layers are denoted with $D'_{\hat{d}, d'} (\cdot)$ and $U'_{d', \hat{d}} (\cdot)$, respectively. 
We obtain output $\vo \in \mathbb{R}^d$ for hierarchical autodencoder $\mathrm{HAE}_{d, \hat{d}, d'}(\cdot)$ as,
\begin{equation}\small
        \begin{split}
            \vo \! = \! \mathrm{HAE}_{d, \hat{d}, d'}(\vx) \!&=\! U(\mathrm{AE}_{\hat{d}, d'}(D(\vx))) \\
            \!&=\! U(U'(D'(D(\vx)))) \\
        \end{split}
\label{eq:hae}
\end{equation}

Note that, HAEs are different from the AEs having multiple encoder and decoder layers, since it also enforces reconstruction loss on the outputs of $D(\vx)$ and $U'(D'(D(\vx)))$ as expressed in Eq.~\ref{eq:hae}. Thus HAEs can compress representations step by step, and provides flexibility to reconstruct the inputs with or without the sub-autoencoders. By sharing the outer layer parameters of a series of inner AEs, HAE can introduce more diversity without adding significant parameters and training overhead.


\paragraph{Stochastic Hierarchical Autoencoder} We use a stochastic set of sub-autoencoders $\mathcal{E}_{\hat{d}}$ within an HAE to formulate a stochastic hierarchical autoencoder, where $\hat{d}$ is the input dimension of the sub-autoencoders.
While performing the forward pass of the stochastic hierarchical autoencoder, we randomly choose one sub-autoencoder $\mathrm{AE}_{\hat{d}, d_i} \in \mathcal{E}_{\hat{d}}$ within the autoencoder $\mathrm{AE}_{d, \hat{d}}$. We set the compression dimension $d_i = \hat{d}/2$. Hence, for an input $\vx \in \mathbb{R}^d$, the output $\vo \in \mathbb{R}^d$ of a stochastic hierarchical autoencoder $\mathrm{HAE}^S_{d, \hat{d}}$ is given by

\begin{equation}\small
    \begin{split}
        \vo  & =  \mathrm{HAE}^S_{d, \hat{d}}(\vx) \\
        & = \begin{cases} 
            U(\mathrm{AE}_{\hat{d},d_i}(D(\vx))), & z > 0.3 \\
            U(D(\vx)), & z \leq 0.3
            \end{cases}
    \end{split}
    \label{eq:shae}
\end{equation}
where $z$ is uniformly sampled from the range $[0, 1]$, and $\mathrm{AE}_{\hat{d},d_i} \in \mathcal{E}_{\hat{d}}$ is randomly selected in each step. So for 30\% of the time, we do not use the sub-autoencoders. These randomness introduces more diversity to the generated views to reduce overfitting. For the stochastic HAE, we only compute the reconstruction loss between $\vx$ and $\vo$, since this also implicitly minimizes the distance between $\mathrm{AE}_{\hat{d},d_i}(D(\vx))$ and $D(\vx)$ in Eq.~\ref{eq:shae}. See \S\ref{sec:appendix_shae_recon_loss} for detailed explanation. In the following sections, we use HAE to denote stochastic hierarchical autoencoder. We also name the hyper-parameter $\hat{d}$ as HAE compression dimension.


\subsection{Multi-View Compressed Representation}\label{subsection:MVCR}

Autoencoders can represent data effectively as compressed vectors that capture the main variability in the data distribution \cite{rumelhart1985learning}.
We leverage this capability of autoencoders within our proposed stochastic networks to formulate \textbf{M}ulti-\textbf{V}iew \textbf{C}ompressed \textbf{R}epresentation (\textbf{MVCR}). 

We give an illustration of MVCR in  Fig.~\ref{fig:model}. We insert \textit{multiple} stochastic HAEs (\cref{subsection:HAE}) between multiple layers of a PLM (shown for one layer in the figure). Following \cref{subsection:compression}, we consider the Transformer-based PLM $f(\cdot)$ with hidden dimension $d$ and $N$ layers. At layer $n$, we denote the set of stochastic HAEs with $\mathrm{HAE}^{S, n} = \{\mathrm{HAE}^{S,n}_{d, \hat{d}_1}, \mathrm{HAE}^{S, n}_{d, \hat{d}_2}, \dots, \mathrm{HAE}^{S, n}_{d, \hat{d}_M}\}$, where $M$ is total the number of HAEs. To prevent discarding useful information in the compression process and for more stable training, we only use them for 50\% of the times. We modify the forward pass of $f(\cdot)$ with MVCR, denoted by $F^{\mathrm{MVCR}}(\cdot, \mathrm{HAE}^{S, n})$ to obtain the output $\vh_n$ for an input $\vh_{n-1}$ as
\begin{equation}\small
    \begin{split}
    \vh_n &= F^{\mathrm{MVCR}}(\vh_{n-1}, \mathrm{HAE}^{S, n}) \\
    &= \begin{cases}
            F^S(f_{n}(\vh_{n-1}), \mathrm{HAE}^{S, n}), & z \leq 0.5 \\
            f_{n}(\vh_{n-1}), & z > 0.5
          \end{cases}
    \end{split}
  \label{eqn:fmvcr}
\end{equation}
where $z$ is uniformly sampled from the range $[0, 1]$. Following Eq.~\ref{eqn:hn}, we finally define $\mathrm{MVCR}(\cdot, \mathrm{HAE}^{S})$ for  a layer set $\mathcal{L}$ of a PLM to compute the output $\vh_n$ using stochastic HAEs as

\begin{equation}\small
    \begin{split}
    \vh_n &= \mathrm{MVCR}(\vh_{n-1}, \mathrm{HAE}^{S}) \\
    &= \begin{cases}
            F^{\mathrm{MVCR}}(f_{n}(\vh_{n-1}), \mathrm{H}^{S, n}), & n \in \mathcal{L} \\
            f_{n}(\vh_{n-1}), & n \notin \mathcal{L}
          \end{cases}
    \end{split}
  \label{eqn:fmvcr2}
\end{equation}
Note that MVCR can be used either at layer-level or token-level. At layer-level, all of the hidden (token) representations in the same layer are augmented with the same randomly selected HAE in each training step. At token-level, each token representation can select a random HAE to use. We show that token-level MVCR performs better than layer-level on account of more diversity and stochasticity (\cref{section:analysis}) and choose this as the default setup. 





\paragraph{Network Optimization} We use two losses for optimizing MVCR. Firstly, our method do not make change to the task layer, so the original task-specific loss $\mathcal{L}_{\text{task}}$ is used to update the PLM, task layer, and HAE parameters. We use a small learning rate $\alpha_{\text{task}}$ to minimize $\mathcal{L}_{\text{task}}$. Secondly, to increase training stability, we also apply reconstruction loss $\mathcal{L}_{\mathrm{MSE}}$ to the output of HAEs to ensure the augmented representations are projected to the space not too far from the original one. At layer $n$, we have
\begin{equation}\small
     \mathcal{L}_{\mathrm{MSE}} = \frac{1}{M\times L}\sum_{m=1}^{M}\sum_{i=1}^{L}(\vh_n^i - \mathrm{HAE}_m(\vh_n^i))^2
     \label{eqn:mseLoss}
\end{equation}
where $M$ is the number of HAEs on each layer, and $L$ is the PLM input sequence length (in tokens). We use a larger learning rate $\alpha_{\mathrm{MSE}}$ to optimize $\mathcal{L}_{\mathrm{MSE}}$ since the HAEs are randomly initialized. This reconstruction loss not only increases training stability, but also allows us to plug-out the HAEs during inference, since it ensures that the generated views are close to the original hidden representation.

\section{Experiments}
In this section, we present the experiments designed to evaluate our method. We first describe the baselines and training details, followed by the evaluation on both the sequence- and token-level tasks across six (6) different datasets.

\paragraph{Baselines}
We compare our method with several widely used parameter and hidden representation regularization methods, including Dropout \citep{dropout}, Weight Decay (WD) \citep{weightdecay} and Gaussian Noise (GN) \citep{vae}.\footnote{We use the same setting as \citep{vibert} to produce these baseline results.}  Some more recent methods are also included for more comprehensive comparison, such as Adapter \citep{adapter}, Mixout \citep{mixout}, as well as the variational information bottleneck (VIB) based method proposed by \citet{vibert}. \footnote{Some of our reproduced VIB results are lower than the results reported in VIB paper though we use the exact same code and hyper-parameters released by the authors. To ensure a fair comparison, we report the average result of the same three random seeds for all experiments.} More information about these baselines can be found in \S\ref{sec:appendix_baselines}. In this paper, we focus on regularization methods for hidden representation. Thus, we do not compare with data augmentation methods which improve model performance from a different aspect.

\begin{table}[t!]
    \centering
    \scalebox{0.54}{
        \begin{tabular}{lcccccc}
        \toprule
        \multicolumn{1}{c|}{\textbf{Method}}& \textbf{full} & \textbf{100} & \textbf{200} & \textbf{500} & \textbf{1000} & \textbf{$\text{avg}_{low}$} \\ \hline
        \multicolumn{7}{c}{SNLI}                                                                                         \\ \hline
        \multicolumn{1}{l|}{BERT}           & 90.28 (0.1)& 48.93 (1.9)       & 58.46 (0.9)       & 66.46 (1.2)       & 72.57 (0.5)        & 61.61         \\
        \multicolumn{1}{l|}{Dropout}        & 90.20 (0.4)& 48.16 (3.4)       & 59.47 (1.3)       & 67.10 (1.8)       & 73.10 (0.3)        & 61.96         \\
        \multicolumn{1}{l|}{WD}             & 90.44 (0.3)& 47.37 (4.6)       & 59.10 (0.3)       & 66.59 (1.3)       & 73.41 (0.7)        & 61.62         \\
        \multicolumn{1}{l|}{GN}             & 90.23 (0.0)& 49.01 (5.6)       & 58.86 (0.6)       & 66.46 (1.9)       & 73.02 (0.3)        & 61.84         \\
        \multicolumn{1}{l|}{Adapter}        & 90.11 (0.0)& 48.88 (1.0)       & 58.99 (0.6)       & 65.80 (0.8)       & 73.67 (0.5)        & 61.84         \\
        \multicolumn{1}{l|}{Mixout}         & 89.97 (0.4)& 49.57 (1.4)       & 57.31 (3.0)       & 63.46 (1.7)       & 71.09 (0.9)        & 60.36         \\
        \multicolumn{1}{l|}{VIB}            & 90.11 (0.1)& 49.46 (1.9)       & 59.14 (1.7)       & 66.21 (0.6)       & 73.87 (1.0)        & 62.17         \\ \hline
        \multicolumn{1}{l|}{MVCR$_{1}$}     & \textbf{90.64 (0.2)}& \textbf{51.73 (0.3)}& \textbf{61.50 (0.1)}& 67.26 (0.1)       & \textbf{74.41 (0.1)}        & \textbf{63.73}\\
        \multicolumn{1}{l|}{MVCR$_{12}$}    & 90.57 (0.2)& 49.99 (1.0)       & 59.45 (0.1)       & \textbf{67.93 (0.4)}       & 73.85 (1.0)        & 62.81             \\ \hline
        \multicolumn{7}{c}{MNLI}                                                                                         \\ \hline
        \multicolumn{1}{l|}{BERT}           & 83.91 (0.5)& 37.27 (4.6)       & 45.60 (1.3)       & 53.22 (2.3)       & 58.58 (1.5)        & 48.67         \\
        \multicolumn{1}{l|}{Dropout}        & 83.94 (0.5)& 37.97 (1.8)       & 43.31 (0.5)       & 52.57 (1.6)       & 59.25 (2.5)        & 48.28         \\
        \multicolumn{1}{l|}{WD}             & 84.27 (0.2)& 38.05 (5.5)       & 45.57 (1.3)       & 52.83 (1.8)       & 58.61 (1.1)        & 48.76         \\
        \multicolumn{1}{l|}{GN}             & 84.00 (0.1)& 38.05 (5.3)       & 45.60 (1.4)       & 53.20 (1.4)       & 58.93 (0.5)        & 48.95         \\
        \multicolumn{1}{l|}{Adapter}        & 83.35 (0.0)& 36.61 (4.5)       & 45.28 (2.9)       & 55.27 (0.3)       & 59.30 (1.9)        & 49.12         \\
        \multicolumn{1}{l|}{Mixout}         & 83.80 (0.2)& 36.61 (4.2)       & 44.92 (1.6)       & 52.89 (0.9)       & 54.88 (1.5)        & 47.33         \\
        \multicolumn{1}{l|}{VIB}            & 84.32 (0.1)& 39.34 (1.8)       & 45.05 (0.9)       & 52.23 (0.2)       & 52.40 (0.3)        & 47.26         \\ \hline
        \multicolumn{1}{l|}{MVCR$_{1}$}     & \textbf{84.47 (0.1)}& \textbf{40.19 (1.1)}& \textbf{46.84 (0.9)}& 56.43 (1.3)   & 59.71 (1.7)        & \textbf{50.79}         \\
        \multicolumn{1}{l|}{MVCR$_{12}$}    & 84.42 (0.2)& 39.61 (2.0)        & 45.94 (1.0)        & \textbf{57.02 (1.6)}& \textbf{60.35 (0.9)}& 50.73        \\ \hline
        \multicolumn{7}{c}{MNLI-mm}                                                                                      \\ \hline
        \multicolumn{1}{l|}{BERT}           & 84.32 (0.6)& 38.69 (3.0)       & 46.20 (1.7)       & 53.44 (2.3)       & 59.88 (1.8)        & 49.55        \\
        \multicolumn{1}{l|}{Dropout}        & 84.44 (0.4)& 38.84 (3.0)       & 44.14 (1.0)       & 53.60 (2.8)       & 60.31 (1.8)        & 49.22        \\
        \multicolumn{1}{l|}{WD}             & 84.57 (0.3)& 39.52 (1.8)       & 46.22 (1.8)       & 53.40 (2.0)       & 60.26 (1.6)        & 49.85        \\
        \multicolumn{1}{l|}{GN}             & 84.33 (0.2)& 40.21 (2.5)       & 47.00 (0.9)       & 53.51 (0.3)       & 60.22 (0.9)        & 50.23        \\
        \multicolumn{1}{l|}{Adapter}        & 84.01 (0.1)& 39.57 (0.8)       & 46.64 (1.7)       & 54.87 (0.9)       & 59.84 (3.2)        & 50.23        \\
        \multicolumn{1}{l|}{Mixout}         & 84.10 (0.2)& 39.88 (1.1)       & 46.22 (1.2)       & 53.48 (0.5)       & 56.09 (0.5)        & 48.92        \\
        \multicolumn{1}{l|}{VIB}            & 84.65 (0.2)& 40.49 (2.2)       & 45.76 (0.2)       & 54.62 (1.7)       & 54.45 (0.7)        & 48.83        \\ \hline
        \multicolumn{1}{l|}{MVCR$_{1}$}     & \textbf{84.72 (0.1)}& 41.75 (0.6)       & \textbf{48.33 (0.9)}& 56.45 (1.2)       & \textbf{61.09 (0.5)}        & 51.90        \\
        \multicolumn{1}{l|}{MVCR$_{12}$}    & 84.68 (0.2)& \textbf{41.82 (0.4)}& 48.32 (1.3)       & \textbf{57.87 (1.5)}       & 60.91 (1.3)        & \textbf{52.23}        \\ \bottomrule
        \end{tabular}
    }
    \caption{Experimental results (Acc.) on the natural language inference (NLI) tasks. \textbf{$\text{avg}_{low}$} is the average result of low-resource scenarios.}
    \label{tb:nli_new}
\end{table}

\begin{table}[t!]
    \centering
    \scalebox{0.54}{
    
    \begin{tabular}{lcccccc}
    \toprule
    \multicolumn{1}{c|}{\textbf{Method}} & \textbf{full}& \textbf{100} & \textbf{200} & \textbf{500} & \textbf{1000} & \textbf{$\text{avg}_{low}$} \\ \hline
    \multicolumn{7}{c}{IMDb}                                                                                         \\ \hline
    \multicolumn{1}{l|}{BERT}           & 88.47 (0.6)& 72.04 (4.8)       & 79.01 (4.3)       & 81.16 (0.8)       & 84.24 (0.5)        & 79.11         \\
    \multicolumn{1}{l|}{Dropout}        & 88.93 (0.1)& 75.89 (2.7)       & 79.76 (2.0)       & 82.19 (0.2)       & 83.85 (0.6)        & 80.42         \\
    \multicolumn{1}{l|}{WD}             & 88.73 (0.6)& 73.74 (4.2)       & 79.61 (3.2)       & 81.90 (0.7)       & 84.00 (0.4)        & 79.81         \\
    \multicolumn{1}{l|}{GN}             & 88.85 (0.1)& 74.17 (2.1)       & 79.49 (2.8)       & 82.10 (0.6)       & 84.04 (0.2)        & 79.95         \\
    \multicolumn{1}{l|}{Adapter}        & 88.39 (0.0)& 73.28 (3.1)       & 80.18 (0.4)       & 81.54 (0.6)       & 83.19 (0.3)        & 79.55         \\
    \multicolumn{1}{l|}{Mixout}         & 88.46 (0.2)& 66.04 (3.0)       & 78.84 (3.8)       & 81.33 (0.8)       & 83.64 (0.3)        & 77.46         \\
    \multicolumn{1}{l|}{VIB}            & 88.73 (0.1)& 73.67 (2.2)       & 79.63 (2.8)       & 82.53 (0.1)       & 83.78 (0.8)        & 79.90         \\ \hline
    \multicolumn{1}{l|}{MVCR$_{1}$}     & \textbf{89.01 (0.1)}& 76.30 (2.1)       & \textbf{81.76 (0.1)}& \textbf{83.22 (0.4)}& \textbf{84.38 (0.2)}         & \textbf{81.41}         \\
    \multicolumn{1}{l|}{MVCR$_{12}$}    & 88.73 (0.1)& \textbf{77.23 (1.8)}& 80.64 (1.3)       & 83.15 (0.3)       & 84.19 (0.4)        & 81.30        \\ \hline
    \multicolumn{7}{c}{Yelp}                                                                                         \\ \hline
    \multicolumn{1}{l|}{BERT}           & 62.01 (0.4)& 40.79 (0.2)       & 42.43 (2.7)       & 46.93 (2.3)     & 51.17 (1.9)        & 45.33             \\
    \multicolumn{1}{l|}{Dropout}        & 62.05 (0.4)& 40.55 (1.3)       & 40.05 (0.4)       & 48.26 (1.8)       & 51.94 (1.3)        & 45.20         \\
    \multicolumn{1}{l|}{WD}             & 61.62 (0.2)& 40.86 (0.2)       & 41.46 (3.5)       & 47.04 (3.1)       & 50.74 (0.8)        & 45.02         \\
    \multicolumn{1}{l|}{GN}             & 62.12 (0.0)& 40.46 (0.8)       & 41.36 (0.5)       & 46.07 (2.5)       & 51.90 (0.9)        & 44.95         \\
    \multicolumn{1}{l|}{Adapter}        & 61.67 (0.3)& 39.58 (0.8)       & 41.05 (0.8)       & 47.49 (1.8)       & 50.26 (0.9)        & 44.60         \\
    \multicolumn{1}{l|}{Mixout}         & 61.50 (0.6)& 40.75 (0.3)       & 40.68 (2.9)       & 45.41 (1.8)       & 49.33 (1.8)        & 44.04         \\
    \multicolumn{1}{l|}{VIB}            & 60.65 (0.4)& 40.84 (0.8)        & 41.21 (0.9)        & 45.59 (2.7)       & 50.22 (3.0)      & 44.46         \\ \hline
    \multicolumn{1}{l|}{MVCR$_{1}$}     & 62.26 (0.3)& \textbf{42.75 (0.2)}&\textbf{43.74 (0.2)}& \textbf{48.50 (0.8)}& 52.68 (0.6)    & \textbf{46.92}        \\
    \multicolumn{1}{l|}{MVCR$_{12}$}    & \textbf{62.42 (0.1)}& 41.83 (1.0)        & 43.10 (0.7)        & 48.02 (1.0)       & \textbf{52.81 (0.4)}& 46.44        \\ \bottomrule
    \end{tabular}

    }
    \caption{Experimental results (Acc.) on the IMDb and Yelp text classification tasks. \textbf{$\text{avg}_{low}$} is the average result of low-resource scenarios.}
    \label{tb:cls_new}
\end{table}

\paragraph{Training Details}
Same as \citet{vibert}, we fine-tune $\text{BERT}_\textrm{base}$ \citep{devlin-etal-2019-bert} for the sequence-level tasks. The token-level tasks are multilingual, so we tune $\text{XLM-R}_\textrm{base}$ \citep{conneau-etal-2020-unsupervised} on them. We use $\text{MVCR}_{l}$ to denote our methods, where $l$ is the PLM layer that we insert HAEs. For example in $\text{MVCR}_{1}$, the HAEs are inserted after the 1st transformer layer. 
On each layer, we insert three HAEs with compressed representation dimensions 128, 256 and 512, respectively. 
All HAEs are discarded during inference. For each experiment, we report the average result of 3 runs.

For each dataset, we randomly sample 100, 200, 500 and 1000 instances from the training set to simulate the low-resource scenarios. The same amount of instances are also sampled from the dev sets for model selection. The full test set is used for evaluation. More details can be found in \S\ref{sec:training_details}.

\subsection{Results of Sequence-Level Tasks}
For the sequence-level tasks, we experiment with two natural language inference (NLI) benchmarks, namely SNLI \citep{bowman-etal-2015-large} and MNLI \citep{williams-etal-2018-broad}, and two text classification tasks, namely IMDb \citep{maas-etal-2011-learning} and Yelp \citep{yelp}. Since MNLI has matched and mismatched versions, we report the results on each of them separately.

We present the experimental results for the NLI tasks in Table~\ref{tb:nli_new}, and text classification tasks in Table~\ref{tb:cls_new}. $\text{MVCR}_{1}$ and $\text{MVCR}_{12}$ are our methods that insert HAEs after the 1st and 12th transformer layer, respectively.\footnote{More details about the hyper-parameter search space for insertion layers can be found in \S\ref{sec:appendix_psearch}.}
We have the following observations: 
\textbf{(a)} $\text{MVCR}_{1}$ and $\text{MVCR}_{12}$ consistently outperform all the baselines in the low-resource settings, demonstrating the effectiveness of our method. 
\textbf{(b)} In \textit{very} low-resource settings such as with 100 and 200 training samples, $\text{MVCR}_{1}$ often performs better than $\text{MVCR}_{12}$. This indicates that bottom-layer hidden representation augmentation is more efficient than top-layer regularization on extreme low-resource data. We attribute this to the fact that the bottom-layer augmentation impacts more parameters on the top layers.
\textbf{(c)} We observe that MVCR outperforms strong baselines, such as VIB. We believe this is due to the fact that MVCR acts as the best-of-both-worlds. Specifically, it acts as a data augmentation module promoting diversity while also acts as a lossy compression module that randomly discards features to prevent overfitting.

\subsection{Results of Token-Level Tasks}
For token-level tasks, we evaluate our methods on WikiAnn \citep{pan-etal-2017-cross} for NER and Universal Dependencies v2.5 \citep{nivre2017universal} for POS tagging. 
Both datasets are multilingual, so we conduct experiments in the zero-shot cross-lingual setting, where all models are fine-tuned on English first and then evaluated on the other languages directly.
    
As shown in Table~\ref{tb:token_new}, our methods are also proven to be useful on the token-level tasks. $\text{MVCR}_{2,12}$ is our method that inserts HAEs after both the 2nd and 12th transformer layers. Comparing the results of $\text{MVCR}_{2}$ and $\text{MVCR}_{2,12}$, we can see adding one more layer of HAEs indeed leads to consistent performance improvement on WikiAnn. However, $\text{MVCR}_{2}$ performs better on the POS tagging task. We also observe that Mixout generally does not perform well when the number of training samples is very small. Detailed analysis about HAE insertion layers is presented in \S\ref{section:analysis}.

\begin{table}[t!]
    \centering
    \scalebox{0.53}{
    
        \begin{tabular}{lcccccc}
        \toprule
        \multicolumn{1}{c|}{\textbf{Method}} & \textbf{full}& \textbf{100} & \textbf{200} & \textbf{500} & \textbf{1000} & \textbf{$\text{avg}_{low}$}  \\ \hline
        \multicolumn{7}{c}{WikiAnn}                                                                                        \\ \hline
        \multicolumn{1}{l|}{XLM-R}      &  59.67 (0.5)& 46.90 (1.0)        & 50.77 (0.4)        & 52.69 (0.3)         & 55.45 (0.4)    & 51.45          \\
        \multicolumn{1}{l|}{Dropout}    &  59.98 (0.4)&  47.30 (0.9)       & 50.09 (0.8)         & 53.20 (0.2)        & 55.10 (0.3)    & 51.42         \\
        \multicolumn{1}{l|}{WD}         &  58.86 (0.6)&  46.86 (1.2)       & 50.46 (0.8)        & 53.28 (1.3)      & 55.29 (0.6)   & 51.47         \\
        \multicolumn{1}{l|}{GN}         &  59.40  (0.6)&  47.82 (0.9)       & 50.94  (0.7)       & 52.89  (0.4)       & 54.29 (0.5)   & 51.48        \\
        \multicolumn{1}{l|}{Adapter}    &  59.31 (0.3) &  45.65 (0.6)       & 48.17 (0.6)        & 51.26  (0.3)       & 54.06 (0.5)   & 49.79        \\
        \multicolumn{1}{l|}{Mixout}     &  59.32 (0.5)&  43.14  (0.7)      & 43.92 (0.9)        & 50.67  (0.5)       & 50.89 (0.3)   & 47.15         \\
        \hline
        \multicolumn{1}{l|}{MVCR$_{2}$}     &  60.33 (0.3)& 47.35 (0.7)        & \textbf{51.43} (0.5)        & 53.93 (0.3)        & 55.72 (0.6)   & 52.11         \\
        \multicolumn{1}{l|}{MVCR$_{2,12}$}  &  \textbf{60.65} (0.4)& \textbf{48.16} (0.7)        & 51.26 (0.7)        & \textbf{54.43} (0.4)       & \textbf{56.10} (0.5)   & \textbf{52.49}         \\ \hline
        \multicolumn{7}{c}{POS}                                                                                          \\ \hline
        \multicolumn{1}{l|}{XLM-R}          &  72.76 (0.2)&  70.52 (0.5)        & 70.87 (0.3)        & 72.23 (0.3)        & 72.37 (0.4)    & 71.50          \\
        \multicolumn{1}{l|}{Dropout}    &  73.03 (0.1)&  70.25 (0.7)       & 71.22 (0.3)        & 72.02 (0.2)        & 72.67 (0.2)    & 71.54          \\
        \multicolumn{1}{l|}{WD}         &  72.76  (0.4)& 70.44 (0.5)        & 71.21 (0.5)        & 72.14 (0.6)        & 72.58 (0.4)    & 71.59          \\
        \multicolumn{1}{l|}{GN}         &  72.96 (0.2)& 70.47 (0.6)        & \textbf{71.47} (0.2)        & 72.48 (0.4)        & 72.43 (0.1)   & 71.70           \\
        \multicolumn{1}{l|}{Adapter}    &  72.70 (0.3)& 70.65 (0.4)        & 71.07 (0.3)        & 72.08 (0.6)        & 71.75 (0.3)   & 71.39               \\
        \multicolumn{1}{l|}{Mixout}     &  72.84 (0.1)&  68.38 (0.8)       &  69.20 (0.4)       & 70.73  (0.2)       & 71.23 (0.6)    & 69.89          \\
        \hline
        \multicolumn{1}{l|}{MVCR$_{2}$}     &  \textbf{73.05} (0.1)& \textbf{71.12}  (0.5)       & 71.33 (0.4)        & \textbf{72.78} (0.3)        &  \textbf{73.13} (0.3)  & \textbf{72.09}           \\
        \multicolumn{1}{l|}{MVCR$_{2,12}$}  &  72.78 (0.2)& 70.93  (0.4)       & 71.25  (0.3)       & 72.49 (0.5)        &  72.56 (0.2)   & 71.81        \\ \bottomrule
        \end{tabular}

    }
    \caption{Zero-shot cross-lingual NER and POS tagging results (F1-score). \textbf{$\text{avg}_{low}$} is the average result of low-resource scenarios.}
    \label{tb:token_new}
\end{table}

    \begin{figure}[t!]
        \centering
        \includegraphics[width=0.9\columnwidth]{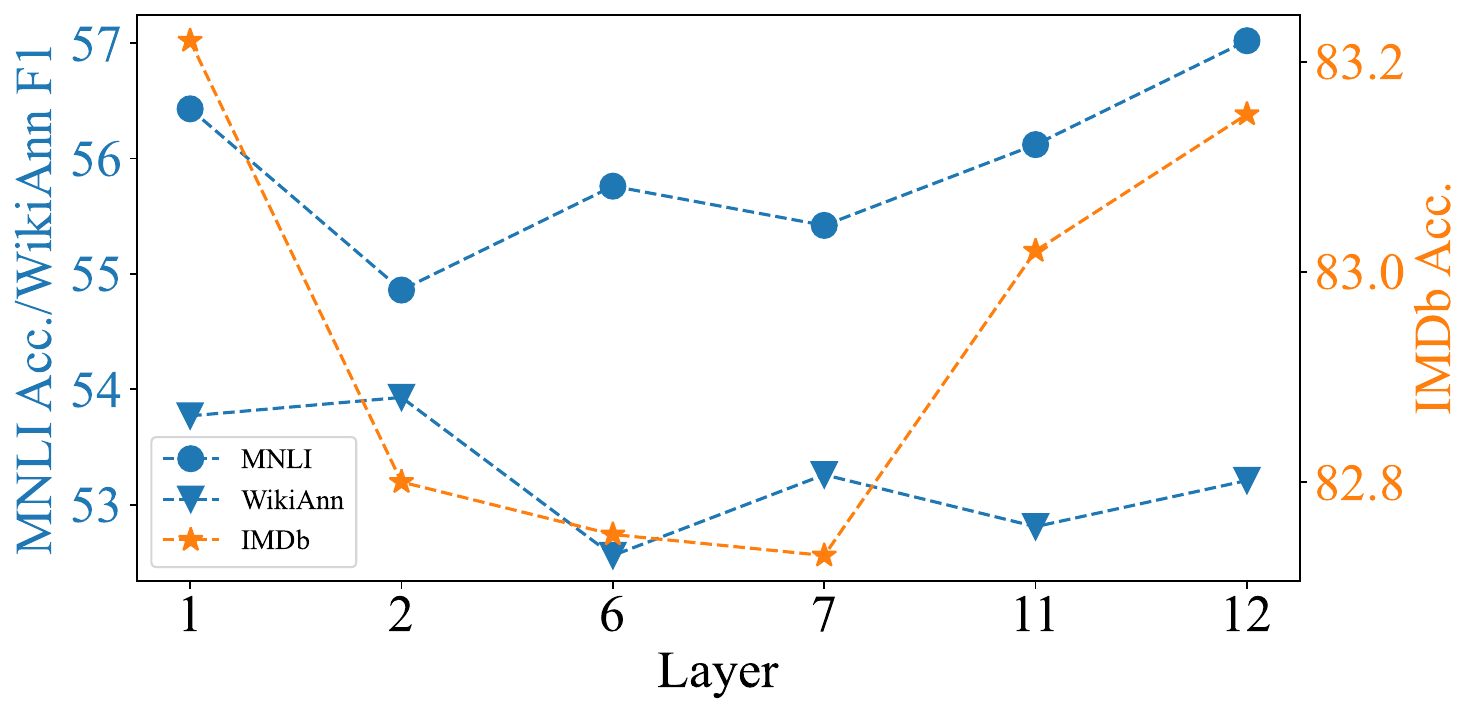}
        \caption{Results on adding the one-layer MVCR after different transformer layers, where bottom and top layers show better results for both sequence- and token-level tasks.}
        \label{fig:ablation_layer}
    \end{figure}

\section{Analysis and Ablation Studies}\label{section:analysis}
To better understand the improvements obtained by MVCR, we conduct in-depth analysis and ablation studies. We experiment with both sequence- and token-level tasks by sampling 500 instances from 3 datasets, namely MNLI, IMDb and WikiAnn.

\paragraph{$\bullet$ Insertion Layer(s) of MVCR}

Based on \citep{clark-etal-2019-bert}, each layer in BERT captures different type of information, such as surface, syntactic or semantic information. Same as above, we fix the dimensions of the three HAEs on each layer to 128, 256 and 512. To comprehensively analyse the impact of adding MVCR after different layers, we select two layers for bottom, middle and top layers. More specifically, \{1, 2\} for bottom, \{6, 7\} for middle  and \{11, 12\} for top layers. 

As shown in Fig.~\ref{fig:ablation_layer}, adding MVCR after layer 1 or 12 achieves the best performance than the other layers. This result can be explained in the way of different augmentation methods \citep{feng-etal-2021-survey}. By adding MVCR after layer 1, the module acts as a data generator which creates multi-view representation inputs to all the upper layers. While adding MVCR after layer 12, which is more relevant to the downstream tasks, the module focuses on preventing the task layer from overfitting \citep{fortuitous_forgetting}.
Apart from one layer, we also experiment with adding MVCR to a combination of two layers (Table~\ref{tb:ablation_layer_sentence} and Table~\ref{tb:ablation_layer_token}). For token-level task, adding MVCR to both layer 2 and 12 performs even better than one layer. However, it does not help improve the performance on sequence-level tasks. Similar to the $\beta$ parameter in VIB \citep{vibert} which controls the amount of random noise added to the representation, the number of layers in MVCR controls the trade-off between adding more variety to the hidden representation and keeping more information from the original representation. And the optimal trade-off point can be different for sequence- and token-level tasks.

\begin{table}[t!]
    \centering
    \scalebox{0.7}{
    \begin{tabular}{lcccccc}
    \toprule
    \textbf{Layer} & \textbf{1} & \textbf{2} & \textbf{6} & \textbf{7} & \textbf{11} & \textbf{12} \\
    \midrule
    MNLI & 56.43 & 54.86 & 55.76 & 55.42 & 56.12 & \textbf{57.02} \\
    IMDb & \textbf{83.22} & 82.80 & 82.75 & 82.73 & 83.02 & 83.15 \\
    \midrule
    \textbf{Layer} & \textbf{1,2} & \textbf{1,3} & \textbf{1,6} & \textbf{1,7} & \textbf{1,11} & \textbf{1,12} \\
    \midrule
    MNLI & 49.94 & 48.63 & 50.38 & 50.73 & 50.52 & \textbf{51.10} \\
    IMDb & 82.59 & \textbf{82.73} & 82.57 & 82.64 & 82.50 & 82.36 \\
    \bottomrule
    \end{tabular}
    }
    \caption{HAE insertion layers for sequence-level tasks.}
    \label{tb:ablation_layer_sentence}
\end{table}

\begin{table}[t!]
    \centering
    \scalebox{0.7}{
    \begin{tabular}{lcccccc}
    \toprule
    \textbf{Layer} & \textbf{1} & \textbf{2} & \textbf{6} & \textbf{7} & \textbf{11} & \textbf{12} \\
    \midrule
    WikiAnn & 53.77 & \textbf{53.93} & 52.56 & 53.26 & 52.81 & 53.21 \\
    \midrule
    \textbf{Layer} & \textbf{2,1} & \textbf{2,3} & \textbf{2,6} & \textbf{2,7} & \textbf{2,11} & \textbf{2,12} \\
    \midrule
    WikiAnn & 53.57 & 53.76 & 53.15 & 54.29 & 53.87 & \textbf{54.43} \\
    \bottomrule
    \end{tabular}
    }
    \caption{HAE insertion layers for token-level task.}
    \label{tb:ablation_layer_token}
\end{table}

\begin{figure}[t]
        \centering
        \includegraphics[width=\columnwidth]{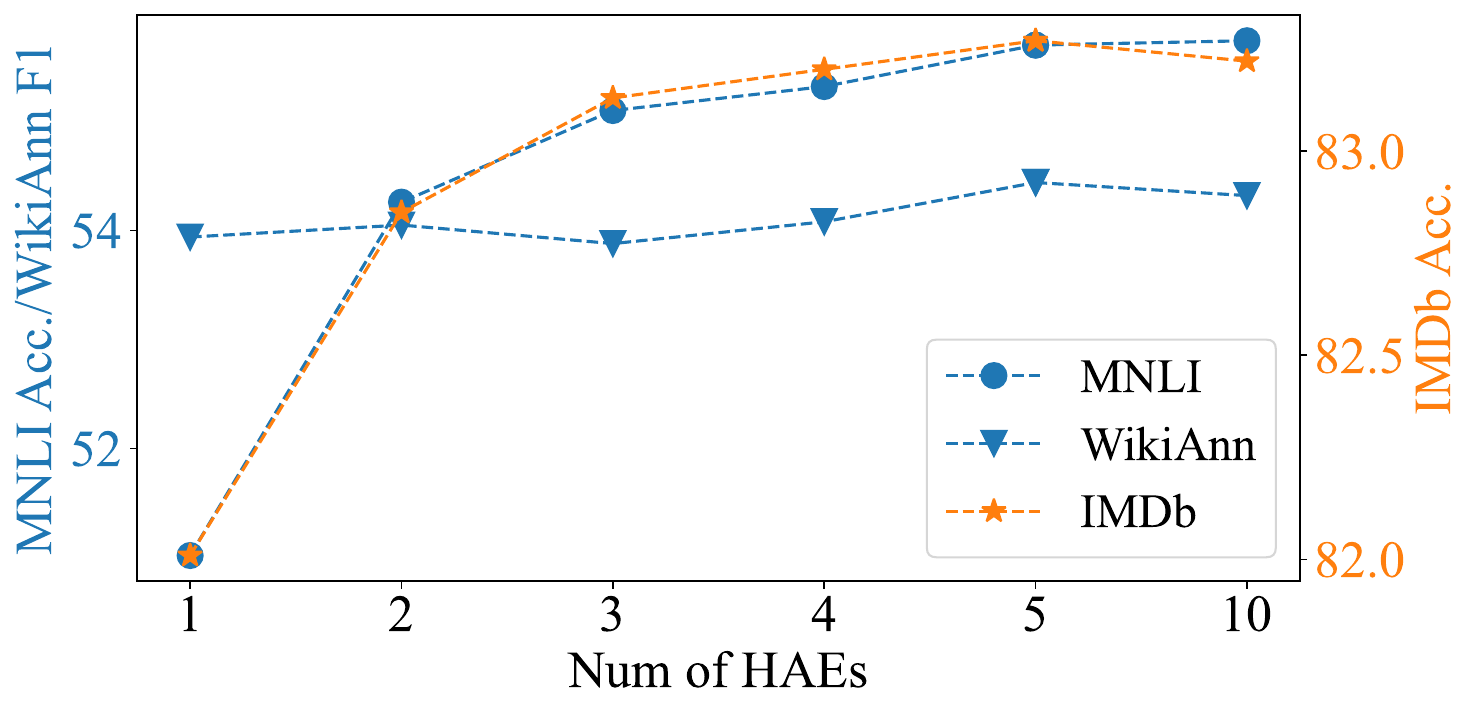}
        \caption{Impact of the number of HAEs in MVCR.}
        \label{fig:ablation_num_aes}
    \end{figure}

\paragraph{$\bullet$ Number of HAEs in MVCR}
We analyse the impact of the number of HAEs in MVCR on the diversity of augmenting hidden representations. We fix the compression dimension of HAE's outer encoder to 256, and only insert HAEs to the bottom layers, layer 1 for MNLI and IMDb, and layer 2 for WikiAnn. As shown in Fig.~\ref{fig:ablation_num_aes}, the performance improves with an increasing number of HAEs, which indicates that adding more HAEs leads to more diversity and better generalization. However, the additional performance gain is marginal after three HAEs. This is probably because without various compressed dimensions, the variety only comes from different initialization of HAEs which can only improve the performance to a limited extent. Hence, we fix the number of HAEs to three for other experiments, including the main experiments.

\begin{figure}[ht]
        \centering
        \includegraphics[width=\columnwidth]{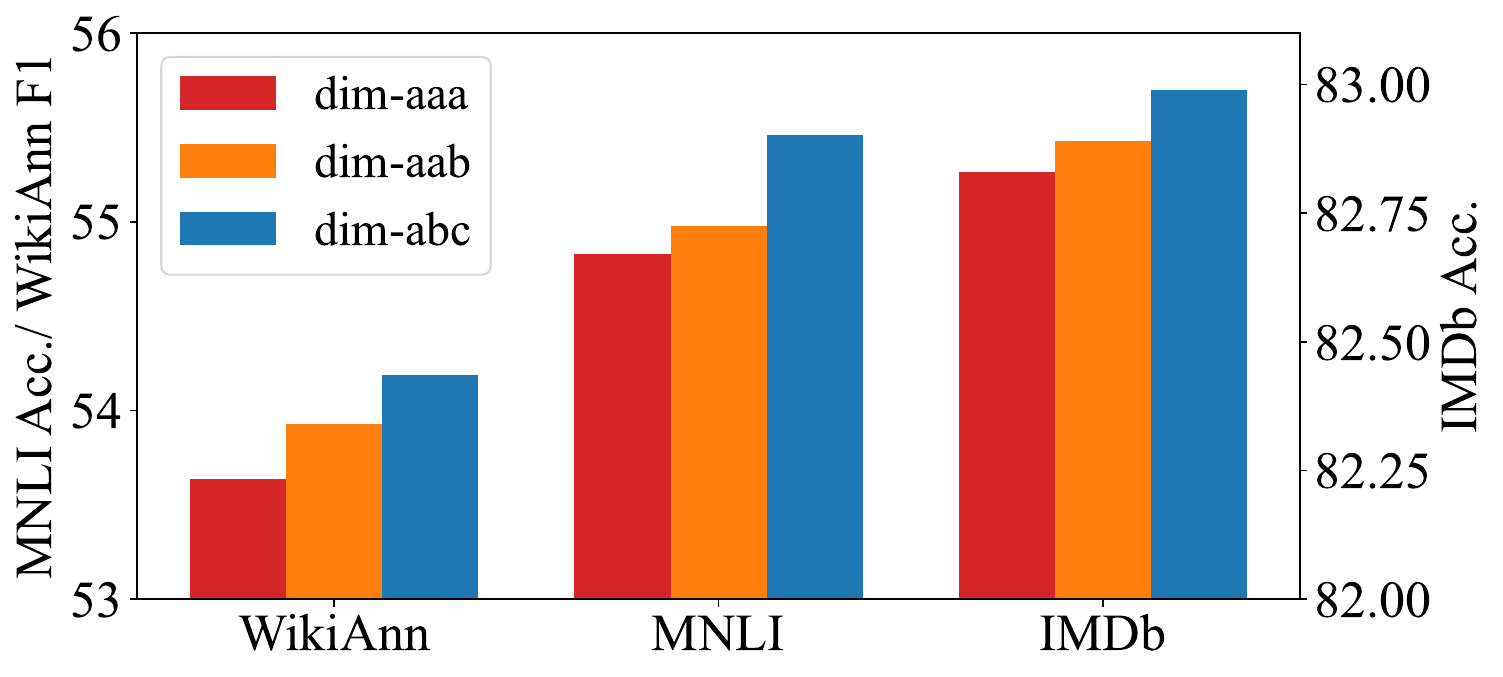}
        \caption{Results on three types of HAE compression dimension combinations: ``aaa'', ``aab'' and ``abc''.}
        \label{fig:ablation_dim}
    \end{figure}

\paragraph{$\bullet$ Diversity of HAE Dimensions}
The compression dimensions of HAEs control the amount of information to be passed to upper PLM layers, so using HAEs of varying dimensions may help generate more diverse views during training.
To analyse the impact of the compression dimension diversity, we run experiments of three types of combinations: ``aaa'', ``aab'' and ``abc'', which contains one, two and three unique dimensions respectively. We compute the average performance of the HAEs with dimensions \{32,32,32\} to \{512,512,512\} for ``aaa''. We sample the dimensions for ``aab'' and ``abc'' since there are too many possible combinations.\footnote{See \S\ref{sec:appendix_stat_results} for more details about the combinations.} As we can see from Fig.~\ref{fig:ablation_dim}, ``abc'' consistently outperforms ``aab'', while ``aab'' consistently outperforms ``aaa'', which indicates increasing compression dimension diversity can help further improve model generalization.

\begin{figure}[ht]
        \centering
            \begin{subfigure}{0.47\columnwidth}
                \centering
                \includegraphics[width=\linewidth]{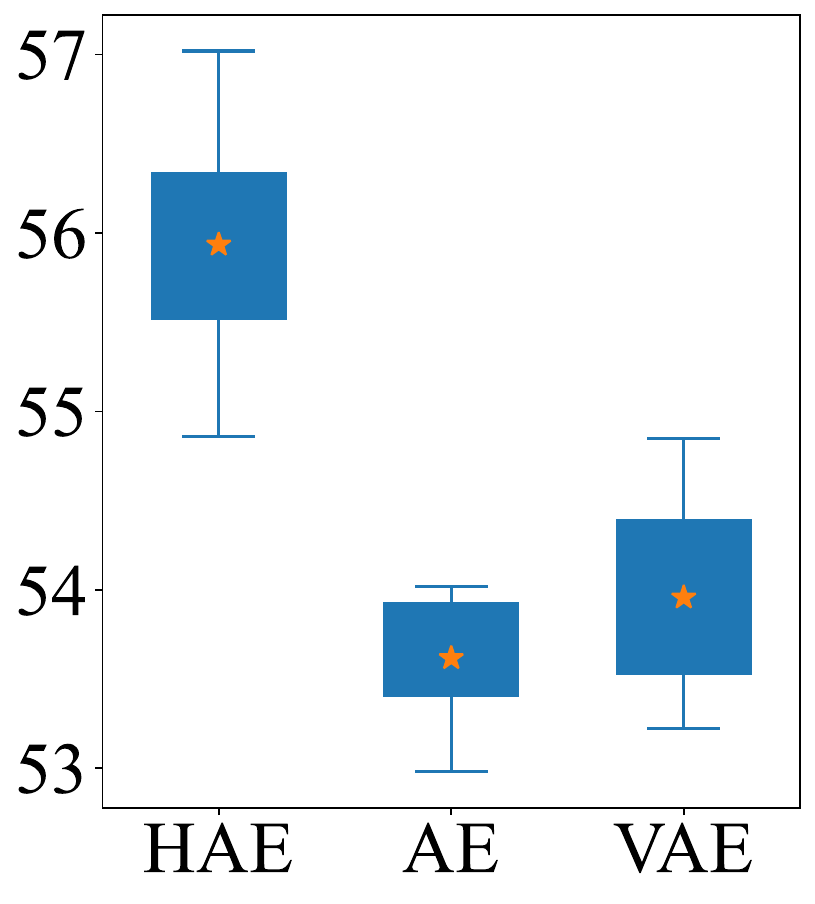}
                \caption{MNLI}
                \label{fig:ablation_ae_type_mnli}
            \end{subfigure}
            \hfill
            \begin{subfigure}{0.5\columnwidth}
                \centering
                \includegraphics[width=\linewidth]{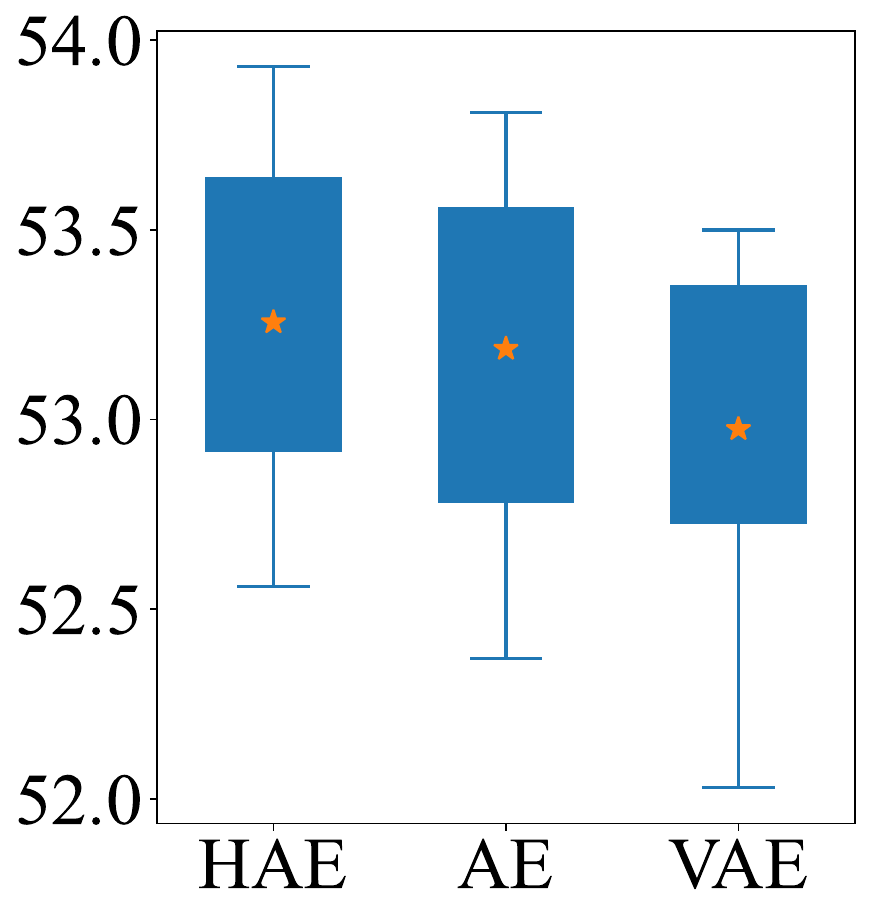}
                \caption{WikiAnn}
                \label{fig:ablation_ae_type_WikiAnn}
            \end{subfigure}
        \caption{Results on different types of autoencoders in MVCR. IMDb results can be found in \S\ref{sec:hae_ae_vae}.}
        \label{fig:ablation_ae_type}
    \end{figure}

\paragraph{$\bullet$ HAE vs. AE vs. VAE}
HAEs in MVCR serve as a bottleneck to generate diverse compressed views of the original hidden representations. There are many other possible alternatives for HAE, so we replace HAE with the vanilla AE and variational autoencoder (VAE) \citep{KingmaW13auto} for comparison. The results in Fig.~\ref{fig:ablation_ae_type} show that HAE consistently outperforms AE and VAE.

\begin{figure}[ht]
        \centering
            \begin{subfigure}{0.48\columnwidth}
                \centering
                \includegraphics[width=\linewidth]{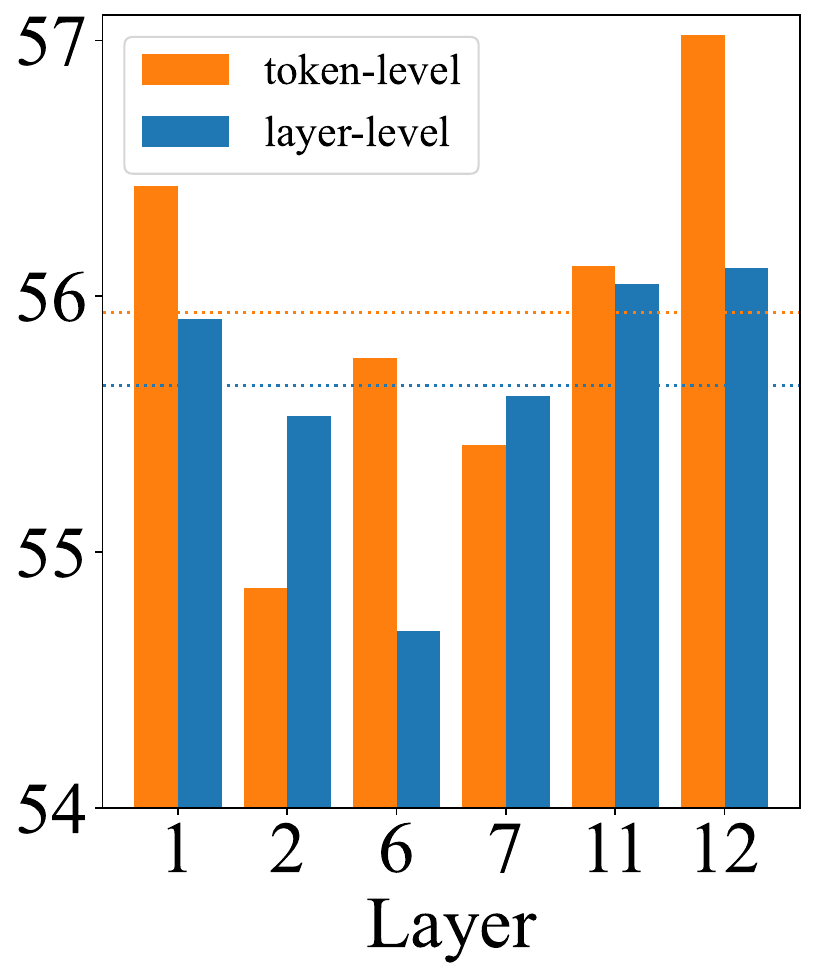}
                \caption{MNLI}
                \label{fig:ablation_layertoken_mnli}
            \end{subfigure}
            \hfill
            \begin{subfigure}{0.48\columnwidth}
                \centering
                \includegraphics[width=\linewidth]{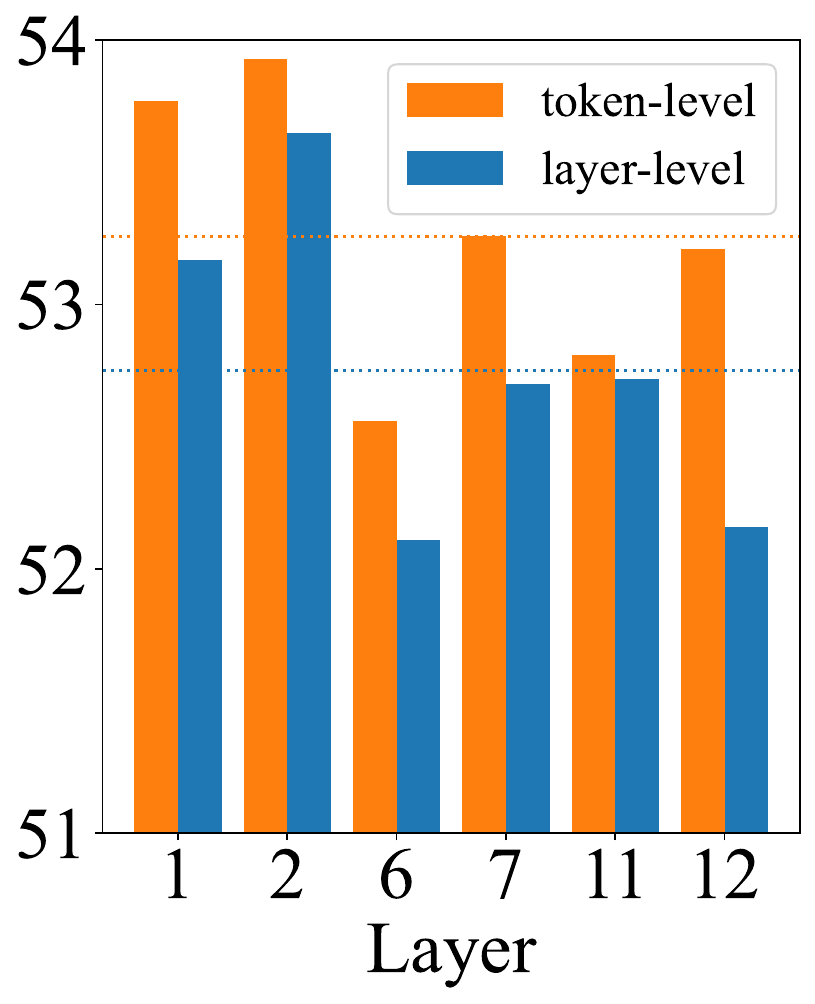}
                \caption{WikiAnn}
                \label{fig:ablation_layertoken_WikiAnn}
            \end{subfigure}
        \caption{Comparison of token- and layer-level MVCRs. IMDb results can be found in \S\ref{sec:hae_ae_vae}.}
        \label{fig:ablation_layertoken}
    \end{figure}

\paragraph{$\bullet$ Token-Level vs. Layer-Level MVCR}
In our method, the selection of random HAE can be on token-level or layer-level. For token-level MVCR, each token in the layer randomly selects an HAE from the pool, with the parameters shared within the same layer. Compared with layer-level MVCR, token-level MVCR adds more variety to the model, leading to better results as observed in Fig.~\ref{fig:ablation_layertoken}.

\paragraph{$\bullet$ More Results and Analysis} We conduct experiments to compare the training overhead of MVCR with other baselines. MVCR has slightly longer training time yet shorter time for convergence. We also experiment with inference with or without MVCR, which results in very close performance. Both experiments can be found in \S\ref{sec:training_overhead}.
\section{Conclusions}

In this work, we have proposed a novel method to improve low-resource fine-tuning robustness via hidden representation augmentation. We insert a set of autoencoders between the layers of a pre-trained language model (PLM). The layers are randomly selected during fine-tuning to generate more diverse compressed representations to prevent the top PLM layers from overfitting. A tailored stochastic hierarchical autoencoder is also proposed to help add more diversity to the augmented representations. The inserted modules are discarded after training, so our method does not add extra parameter or computation cost during inference. Our method has demonstrated consistent performance improvement on a wide range of NLP tasks.
\section*{Limitations}

We focus on augmenting the hidden representations of a PLM.  Thus most of our baselines, such as dropout \citep{dropout} and variational information bottleneck methods \citep{vibert}, do not require unlabeled data. For a fair comparison, we assume that the unlabeled data is not available. Therefore, only the limited labeled training set are used to train the autoencoders in our experiments. However, such unlabeled general- or in-domain data (e.g., Wikipedia text) are easy to obtain in practice, and can be used to pre-train the autoencoders with unsupervised language modeling tasks, which may help further improve the performance. We leave it for future work.
\section*{Ethical Impact}
Deep learning has demonstrated encouraging performance on a wide range of tasks during the past few years. However, neural models are data hungry, which usually requires a large amount of training data to achieve reasonable performance. It is expensive and time consuming to annotate a large amount of data. Pretrained language models (PLMs) \citep{devlin-etal-2019-bert,conneau2019cross,liu-etal-2020-multilingual-denoising} have been proven to be useful to transfer knowledge from massive unlabeled text to downstream tasks, but they are also prone to overfitting during fine-tuning due to over-parameterization. In this work, we propose a novel method to help improve model robustness in the low-resource scenarios, which is part of the attempt to reduce neural model reliance on the labeled data, and hence reduce annotation cost. Our method has also demonstrated promising performance improvement on cross-lingual NLP tasks, which is also an attempt to break the language barrier and allow a larger amount of population to benefit from the advance of NLP techniques.

\section*{Acknowledgements}

This research is supported, in part, by Alibaba Group through Alibaba Innovative Research (AIR) Program and Alibaba-NTU Singapore Joint Research Institute (JRI), Nanyang Technological University, Singapore.

\bibliography{custom}
\bibliographystyle{emnlp22/acl_natbib}

\appendix

\section{Appendix}
\label{sec:appendix}

\subsection{Dataset Usage}
SNLI and Universal Dependencies v2.5 are under Attribution-ShareAlike 4.0 International license, which is free to share and adapt. MNLI is freely available for typical machine learning uses, and may be modified and redistributed. IMDb and Yelp are available for limited non-commercial usage. WikiAnn is under ODC-By license for research usage.

\subsection{Stochastic Hierarchical Autoencoder Implicit Reconstruction Loss}
\label{sec:appendix_shae_recon_loss}
The stochastic hierarchical autoencoder implicitly minimizes the reconstruction loss of its sub-autoencoder inputs and outputs. We can rewrite Eq.~\ref{eq:shae} as $\vo = U(\vx^\prime)$, where 
\begin{equation}\small
    \vx^\prime =  \begin{cases}
            \mathrm{AE}_{\hat{d},d_i}(D(\vx)), & z > 0.3 \\
            D(\vx), & z \leq 0.3
            \end{cases}
  \label{eqn:shae_rewrite}
\end{equation}
Therefore, minimizing the distance between $\vx$ and $\vo$ also enforces the generated $\vx^\prime$ to be as close as possible even when $\vx$ is fed into different neural modules. That implicitly minimizes the distances between $\mathrm{AE}_{\hat{d},d_i}(D(\vx))$ and $D(\vx)$. Let $x^{\prime\prime} = D(\vx)$, so it turns out that the distance between $\mathrm{AE}_{\hat{d},d_i}(x^{\prime\prime})$ and $x^{\prime\prime}$ is also minimized. Therefore, we can omit the explicit reconstruction loss between $\mathrm{AE}_{\hat{d},d_i}(x^{\prime\prime})$ and $x^{\prime\prime}$ during model training, which helps reduce the complexity of our method.

\subsection{Baselines}
\label{sec:appendix_baselines}
In this section, we describe more details about our baseline methods.

\paragraph{Dropout}
Dropout \citep{dropout} regularizes the model by randomly ignoring layer outputs with probability $p$ during training. It has been widely used in almost all PLMs \citep{devlin-etal-2019-bert, roberta, deberta} to prevent overfitting. We use dropout at all layers and choose the best performance from $p\in \{0.25, 0.5, 0.75\}$ as this baseline.

\paragraph{Weight Decay (WD)}
WD \citep{weightdecay} is a widely used regularization technique by adding a term $\frac{\lambda}{2}\|w\|^2$ in the loss function to penalise the size of the weight vector. We adopt a variation of WD and replace the term with $\frac{\lambda}{2}\|w-w_0\|^2$, which is more tailored to the fine-tuning of PLMs \citep{mixout}. We choose the best performance from $\lambda \in \{0.25, 0.5, 0.75\}$ as this baseline.

\paragraph{Gaussian Noise (GN)}
Variational Auto-Encoder (VAE) \citep{vae} shows that adding GN to the hidden layer can improve the model training. We add a small amount of GN, $0.002 \cdot \mathcal{N}(0,I)$, to the hidden output of each layer in PLM during training as this baseline.

\paragraph{Adapter}
Adapter \citep{adapter} is an encoder-decoder module for parameter-efficient transfer-learning in NLP. It freezes the PLM parameters, and only adjust the parameters of the light-weight modules inserted between transformer layers during fine-tuning, which can also be viewed as a regularization method. So we include it in the baselines as a reference.

\paragraph{Mixout}
Motivated by dropout, mixout \citep{mixout} stochastically mixes the parameters of two models, which aims to minimize the deviation of the two. During training, we replace layer outputs with the corresponding values from the initial PLM checkpoint with probability $p$. We choose the best performance from $p \in \{10^{-1}, 10^{-2}, 10^{-3}\}$ as this baseline.

\paragraph{Variational Information Bottleneck (VIB)}
\citet{vibert} leverages VIB \citep{alemi2016deep} to suppress irrelevant features when fine-tuning the model on target tasks. It compresses the sentence representation $\vx$ generated by PLMs into a smaller dimension representation $\vz$ with mean $\mu(\vx)$ and also introduces more diversity through GN with variance $\text{Var}(\vx)$. 

\subsection{Layer-level MVCR}
\begin{figure*}[t!]
    \centering
    \includegraphics[width=\textwidth]{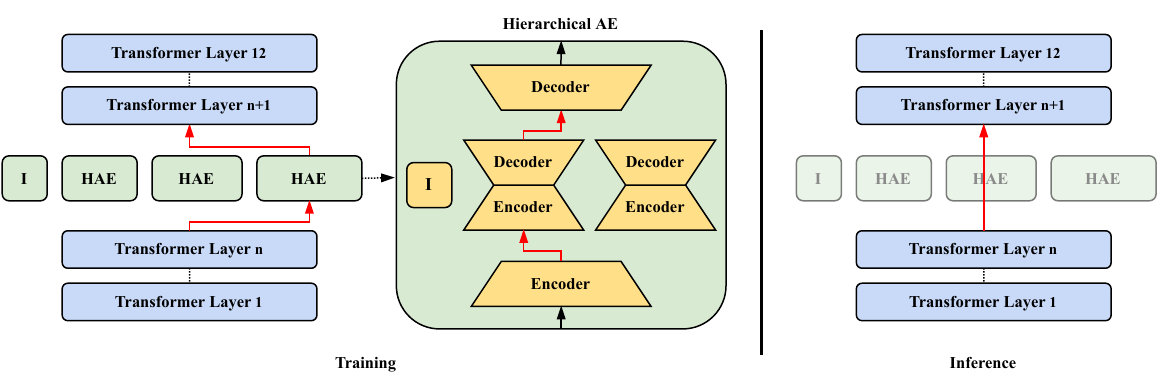}
    \caption{Layer-level MVCR.}
    \label{fig:model_layer}
\end{figure*}

The default setting of MVCR is token-level selection, which means at each training step each token in the same layer randomly selects a different HAE from the pool where the weights are shared. A variation of token-level MVCR is layer-level MVCR (Fig.~\ref{fig:model_layer}), where all tokens in the same layer randomly selects the same HAE from the pool. 

\subsection{Training Details}
\label{sec:training_details}
We use the same hyper-parameters as \citep{vibert} and \citep{hu2020xtreme} in the experiments. The model parameters are optimized with Adam \citep{adam}. We use learning rate 2e-5 ($\alpha_\textrm{task}$ in \S\ref{subsection:MVCR}) to tune all parameters to minimize the task-specific losses, and use 2e-3 ($\alpha_\textrm{MSE}$ in \S\ref{subsection:MVCR}) to tune the HAE parameters to minimize the reconstruction losses. 
For all low-resource experiments, we train the model for 100 epochs with batch size 32. In the first 20 epochs, we freeze the parameters of the PLM and the classifier, and pretrain the HAEs only with the reconstruction loss described in Eq. \ref{eqn:mseLoss}. Then all parameters are tuned in the remaining 80 epochs. All the baseline models are fine-tuned for 80 epochs for a fair comparison with MVCR. Similarly, for full data setting, we train our model for 25 epochs, including 5-epoch HAE pretraining and 20-epoch all-parameter fine-tuning. Similarly, we train the baseline models for 20 epochs. The number of training epochs are sufficiently large for both baseline and our methods to converge. All experiments are conducted on the Nvidia Tesla v100 16GB GPUs, and each low-resource experiment takes about 20 to 40 minutes to complete.

\subsection{Hyper-Parameter Search}
\label{sec:appendix_psearch}
Most of the hyper-parameters we use in the downstream tasks are same as \citep{vibert} and \citep{hu2020xtreme}. Therefore, we only need to decide the hyper-parameters specific to our methods. To reduce the complexity, we only sample 500 samples from the MNLI, IMDb and WikiAnn datasets for hyper-parameters search. Instead of using a different set of hyper-parameters for each dataset, we use the set of hyper-parameters for all of the sentence-level tasks and another set for the token-level tasks. To determine the HAE insertion layers, we conduct hyper-parameter search from 5 choices \{\{1\}, \{2\}, \{12\}, \{1,12\}, \{2,12\}\} with the sampled data, and decide to use \{\{1\}, \{12\}\} for all of the sentence-level tasks, and use 
 \{\{2\}, \{2,12\}\} for all of the token-level tasks. We also conduct more comprehensive analysis of the insertion layers in \S\ref{section:analysis}.

\subsection{More Results and Analysis}
\label{sec:appendix_stat_results}

\paragraph{Number of HAEs in MVCR}
Table \ref{tb:ablation_ae_num} shows the results of adding different number of HAEs to MVCR.
\begin{table}[htbp]
    \centering
    \scalebox{0.8}{
    \begin{tabular}{lcccccc}
    \toprule
    \textbf{N} & \textbf{1} & \textbf{2} & \textbf{3} & \textbf{4} & \textbf{5} & \textbf{10} \\
    \midrule
    MNLI & 51.02 & 54.26 & 55.10 & 55.32 & 55.70 & \textbf{55.74} \\
    IMDb & 82.01 & 83.18 & 83.13 & 83.20 & \textbf{83.27} & 83.22 \\
    WikiAnn & 53.94 & 54.05 & 53.88 & 54.08 & \textbf{54.44} & 54.32 \\
    
    \bottomrule
    \end{tabular}
    }
    \caption{Comparision of numbers of HAEs in MVCR \textbf{N} * \{256\}.}
    \label{tb:ablation_ae_num}
\end{table}

\begin{table*}[t!]
    \centering
    \scalebox{0.8}{
    \begin{tabular}{lccccccc}
    \toprule
    \textbf{AE dim} & \textbf{32,32,32} & \textbf{64,64,64} & \textbf{128,128,128} & \textbf{256,256,256} & \textbf{512,512,512} & \textbf{avg.}\\
    \midrule
    MNLI & 55.08 & 55.51 & 54.85 & 55.10 & 53.62 & 54.83 \\
    IMDb & 82.62 & 82.71 & 82.64 & 83.13 & 83.04 & 82.83 \\
    WikiAnn & 53.41 & 53.69 & 53.77 & 53.88 & 53.43 & 53.64 \\

    \toprule
    \textbf{AE dim} & \textbf{64,64,32} & \textbf{64,64,64} & \textbf{64,64,128} & \textbf{64,64,256} & \textbf{64,64,512} & \textbf{avg.} \\
    \midrule
    MNLI & 56.61 & 55.51 & 55.17 & 54.80 & 52.14 & 54.85 \\
    IMDb & 82.95 & 82.71 & 82.69 & 82.63 & 83.41 & 82.88 \\ 

    \toprule
    \textbf{AE dim} & \textbf{256,256,32} & \textbf{256,256,64} & \textbf{256,256,128} & \textbf{256,256,256} & \textbf{256,256,512} & \textbf{avg.} \\
    \midrule
    MNLI & 54.34 & 55.33 & 56.03 & 55.10 & 54.11 & 54.98 \\
    IMDb & 83.06 & 82.82 & 83.04 & 83.13 & 82.39 & 82.89 \\
    WikiAnn & 53.89 & 54.05 & 54.10 & 53.88 & 53.73 & 53.93 \\

    \toprule
    \textbf{AE dim} & \textbf{128,256,32} & \textbf{128,256,64} & \textbf{128,256,128} & \textbf{128,256,256} & \textbf{128,256,512} & \textbf{avg.}\\
    \midrule
    MNLI & 55.10 & 55.70 & 54.06 & 56.03 & 56.43 & 55.46 \\
    IMDb & 82.99 & 82.85 & 82.87 & 83.04 & 83.22 & 82.99 \\
    WikiAnn & 53.85 & 54.55 & 54.56 & 54.10 & 53.93 & 54.19 \\
    
    \bottomrule
    \end{tabular}
    }
    \caption{Comparision of dimensions of AutoEncoder.}
    \label{tb:ablation_ae_dim}
\end{table*}

\paragraph{Diversity of HAE Dimensions}
We conduct extensive experiments to study the impacts of different types of dimension combinations, including ``aaa'', ``aab'' and ``abc''. Results are presented in Table \ref{tb:ablation_ae_dim}. On avergage, "dim-abc" outperforms the other two types, while "dim-aab" outperforms "dim-aaa". Furthermore, as shown in the first row of Table \ref{tb:ablation_ae_dim}, the results are close for ``aaa'' ranging from \{32,32,32\} to \{512,512,512\}. As such, we choose \{64, 64, x\}, \{256, 256, x\} for the analysis of ``aab'' and \{128,256,x\} for ``abc''. 

\begin{figure}[ht]
        \centering               \includegraphics[width=0.6\linewidth]{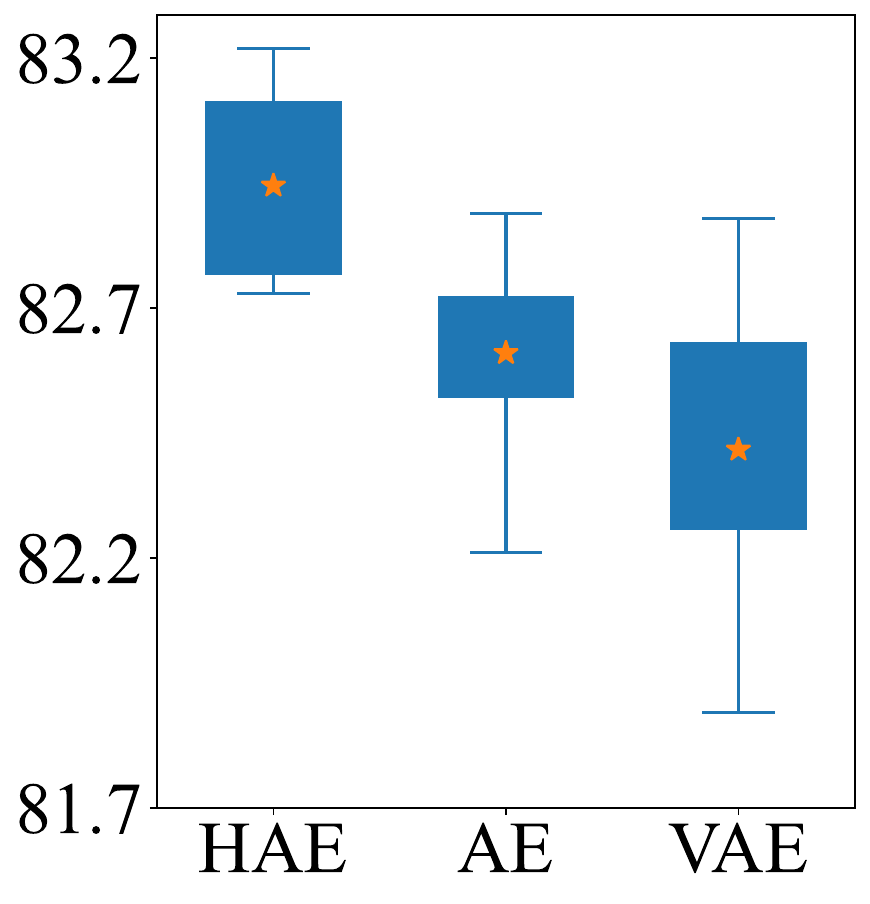}
        \caption{Results on different types of autoencoders in MVCR for IMDb.}
        \label{fig:ablation_ae_type_imdb}
    \end{figure}

\paragraph{HAE vs. AE vs. VAE}
\label{sec:hae_ae_vae}
We replace HAE in MVCR with AE and VAE and compare the results in Table \ref{tb:ablation_ae_vae}. The results for IMDb is also plotted in Fig.~\ref{fig:ablation_ae_type_imdb}.
\begin{table}[htbp]
    \centering
    \scalebox{0.6}{
    \begin{tabular}{lccccccc}
    \toprule
    \textbf{Layer} & \textbf{1} & \textbf{2} & \textbf{6} & \textbf{7} & \textbf{11} & \textbf{12} & \textbf{avg.} \\
    \midrule
    MNLI-AE     & 56.43 & 54.86 & 55.76 & 55.42 & 56.12 & 57.02 & \textbf{55.94} \\
    MNLI-AE (orig) & 52.98 & 53.95 & 53.90 & 54.02 & 53.36 & 53.49 & 53.62 \\
    MNLI-VAE    & 53.22 & 54.07 & 54.52 & 53.49 & 54.85 & 53.59 & 53.96 \\
    \midrule
    IMDb-AE     & 83.22 & 82.80 & 82.75 & 82.73 & 83.02 & 83.15 & \textbf{82.95} \\
    IMDb-AE (orig) & 82.89 & 82.63 & 82.72 & 82.21 & 82.73 & 82.48 & 82.61 \\
    IMDb-VAE & 82.47 & 82.88 & 82.22 & 82.69 & 82.35 & 81.89 & 82.42 \\
    \midrule
    WikiAnn-AE & 53.77 & 53.93 & 52.56 & 53.26 & 52.81 & 53.21 & \textbf{53.26} \\
    WikiAnn-AE (orig) & 53.52 & 53.81 & 52.37 & 53.58 & 53.21 & 52.63 & 53.19 \\
    WikiAnn-VAE & 53.11 & 53.50 & 52.03 & 53.21 & 52.59 & 53.41 & 52.97 \\
    \bottomrule
    \end{tabular}
    }
    \caption{Comparison of AE, VAE and Hierarchical AE.}
    \label{tb:ablation_ae_vae}
\end{table}

\begin{figure}[th!]
        \centering               \includegraphics[width=0.5\linewidth]{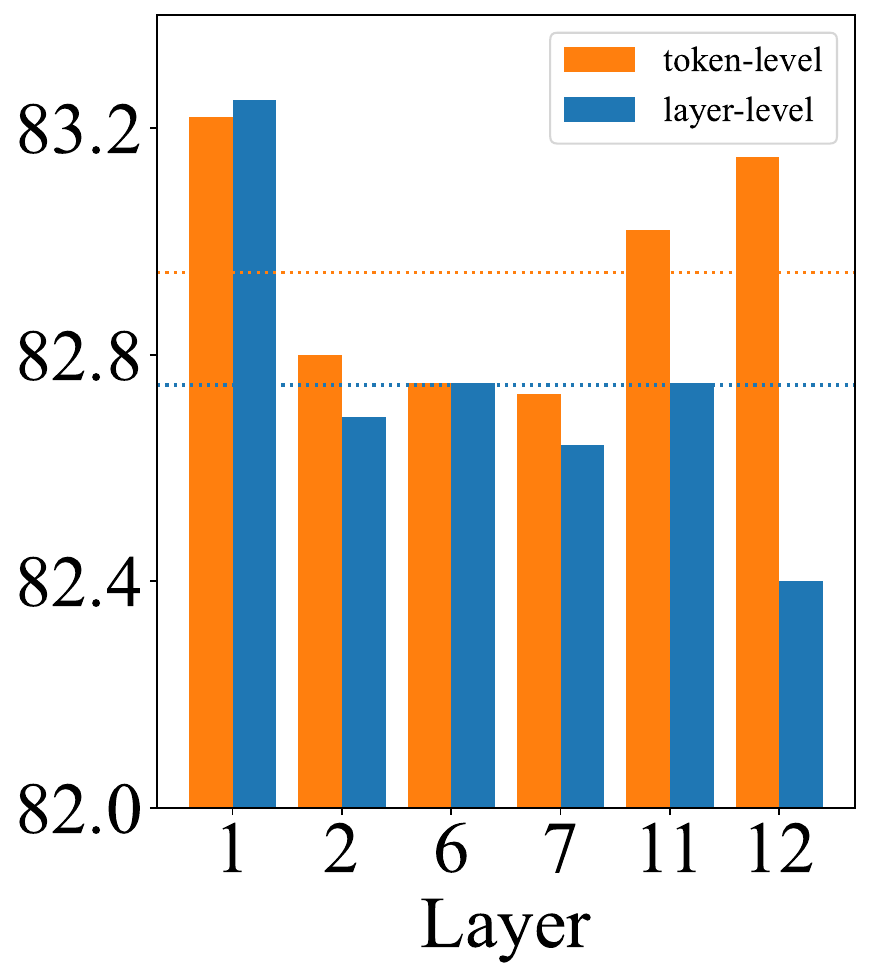}
        \caption{Performance of token- and layer-level MVCR on IMDb.}
        \label{fig:ablation_layertoken_imdb}
    \end{figure}

\paragraph{Token- vs Layer-level MVCR}
\label{sec:token_layer}
We compare the performance of Token- and Layer-level MVCR, and report detailed results in Table~\ref{tb:ablation_tk_layer}. The results for IMDb is also plotted in Fig.~\ref{fig:ablation_layertoken_imdb}.
\begin{table}[htbp]
    \centering
    \scalebox{0.6}{
    \begin{tabular}{lccccccc}
    \toprule
    \textbf{Layer} & \textbf{1} & \textbf{2} & \textbf{6} & \textbf{7} & \textbf{11} & \textbf{12} & \textbf{avg.} \\
    \midrule
    MNLI-Layer & 55.91 & 55.53 & 54.69 & 55.61 & 56.05 & 56.11 & 55.65 \\
    MNLI-TK    & 56.43 & 54.86 & 55.76 & 55.42 & 56.12 & 57.02 & \textbf{55.94} \\
    \midrule
    IMDb-Layer & 83.25 & 82.69 & 82.75 & 82.64 & 82.75 & 82.40 & 82.74 \\
    IMDb-TK & 83.22 & 82.80 & 82.75 & 82.73 & 83.02 & 83.15 & \textbf{82.95} \\
    \midrule
    WikiAnn-Layer & 53.17 & 53.65 & 52.11 & 52.70 & 52.72 & 52.16 & 52.75 \\
    WikiAnn-TK & 53.77 & 53.93 & 52.56 & 53.26 & 52.81 & 53.21 & \textbf{53.26} \\
    \bottomrule
    \end{tabular}
    }
    \caption{Comparison of token- and layer-level MVCR.}
    \label{tb:ablation_tk_layer}
\end{table}

\paragraph{Training Overhead}
\label{sec:training_overhead}
We compare the training runtime and converge time on MNLI with 500 training data in Table~\ref{tb:cost}. The total runtime of 80 epochs for MVCR is only 3.07\% more than the average runtime of all baselines. Moreover, MVCR converges faster than other baselines.
\begin{table}[t!]
    \centering
    \scalebox{0.7}{
    \begin{tabular}{lcccc}
    \toprule
     & \textbf{80 epochs runtime} & \textbf{training converge} & \textbf{converge time} \\
    \midrule
    BERT        & 34m27s & 13 & 335.89s \\
    Dropout     & 34m26s & 21 & 542.33s \\
    Mixout      & 34m48s & 17 & 443.70s \\
    VIB         & 34m44s & 18 & 468.90s \\
    MVCR        & 35m37s & 15 & 400.69s \\
    
    \bottomrule
    \end{tabular}
    }
    \caption{Training time and converge time on MNLI with 500 training data.}
    \label{tb:cost}
\end{table}

\paragraph{Inference With or Without MVCR}
\label{sec:w_wo_mvcr}
We compare the results of inference with or without MVCR on MNLI and IMDb. The results in Table~\ref{tb:hae_inference} shows that the performances are similar.

\begin{table}[t!]
    \centering
    \scalebox{0.7}{
    \begin{tabular}{lccccccc}
    \toprule
    \textbf{Layer} & \textbf{1} & \textbf{2} & \textbf{6} & \textbf{7} & \textbf{11} & \textbf{12} & \textbf{avg.}\\
    \midrule
    MNLI$_{w/o}$ & 56.43 & 54.86 & 55.76 & 55.42 & 56.12 & 57.02 & 55.94 \\
    MNLI$_{w}$ & 55.76 & 56.30 & 56.25 & 54.59 & 56.34 & 56.30 & 55.92 \\
    IMDb$_{w/o}$ & 83.22 & 82.80 & 82.75 & 82.73 & 83.02 & 83.15 & 82.95 \\
    IMDb$_{w}$ & 83.06 & 83.04 & 83.00 & 82.97 & 83.17 & 83.24 & 83.08 \\
    
    \bottomrule
    \end{tabular}
    }
    \caption{If use HAE during inference time.}
    \label{tb:hae_inference}
\end{table}

\end{document}